\documentclass[10pt,twocolumn,letterpaper]{article}

\usepackage{cvpr}              
\usepackage{dsfont}

\definecolor{cvprblue}{rgb}{0.21,0.49,0.74}
\usepackage[pagebackref,breaklinks,colorlinks,allcolors=cvprblue]{hyperref}

\usepackage[dvipsnames]{xcolor}
\usepackage{xspace}

\usepackage{bbm}
\usepackage{booktabs}
\usepackage{graphicx}
\usepackage{multirow}
\usepackage{wrapfig}
\usepackage{colortbl} 
\usepackage[⟨options⟩]{overpic}
\usepackage{stmaryrd}
\usepackage{comment}
\usepackage[normalem]{ulem}

\newcommand{\TW}[1]{{\bf\textcolor{blue}{\bf TW: #1}}}
\newcommand{\CJ}[1]{{\color{magenta} \small(CJ: #1)}}

\newcommand{\method}{\emph{CrossSDF}\xspace}

\definecolor{best}{RGB}{153, 255, 153}    
\definecolor{second}{RGB}{255, 255, 153}  
\makeatletter
\renewcommand{\paragraph}{%
    \@startsection{paragraph}{4}%
    {\z@}{-0.5em}{-0.5em}%
    {\normalfont\normalsize\bfseries}%
}
\makeatother

\usepackage{pifont}

\definecolor{cvprblue}{rgb}{0.21,0.49,0.74}
\usepackage[pagebackref,breaklinks,colorlinks,allcolors=cvprblue]{hyperref}

\def\confName{CVPR}
\def\confYear{2025}

\title{CrossSDF: 3D Reconstruction of Thin Structures From Cross-Sections}

\author{
    Thomas Walker\textsuperscript{1}\footnotemark[1] \and
    Salvatore Esposito\textsuperscript{1}\footnotemark[1] \and
    Daniel Rebain\textsuperscript{2} \and
    Amir Vaxman\textsuperscript{1} \and
    Arno Onken\textsuperscript{1} \and
    Changjian Li\textsuperscript{1} \and
    Oisin Mac Aodha\textsuperscript{1}\and
    \textsuperscript{1}University of Edinburgh \quad
    \textsuperscript{2}University of British Columbia\\
}
\begin{document}

\twocolumn[{%
\maketitle

\thispagestyle{empty}

\begin{center}
\vspace{-7mm}
\captionsetup{type=figure}
\includegraphics[width=.97\linewidth]{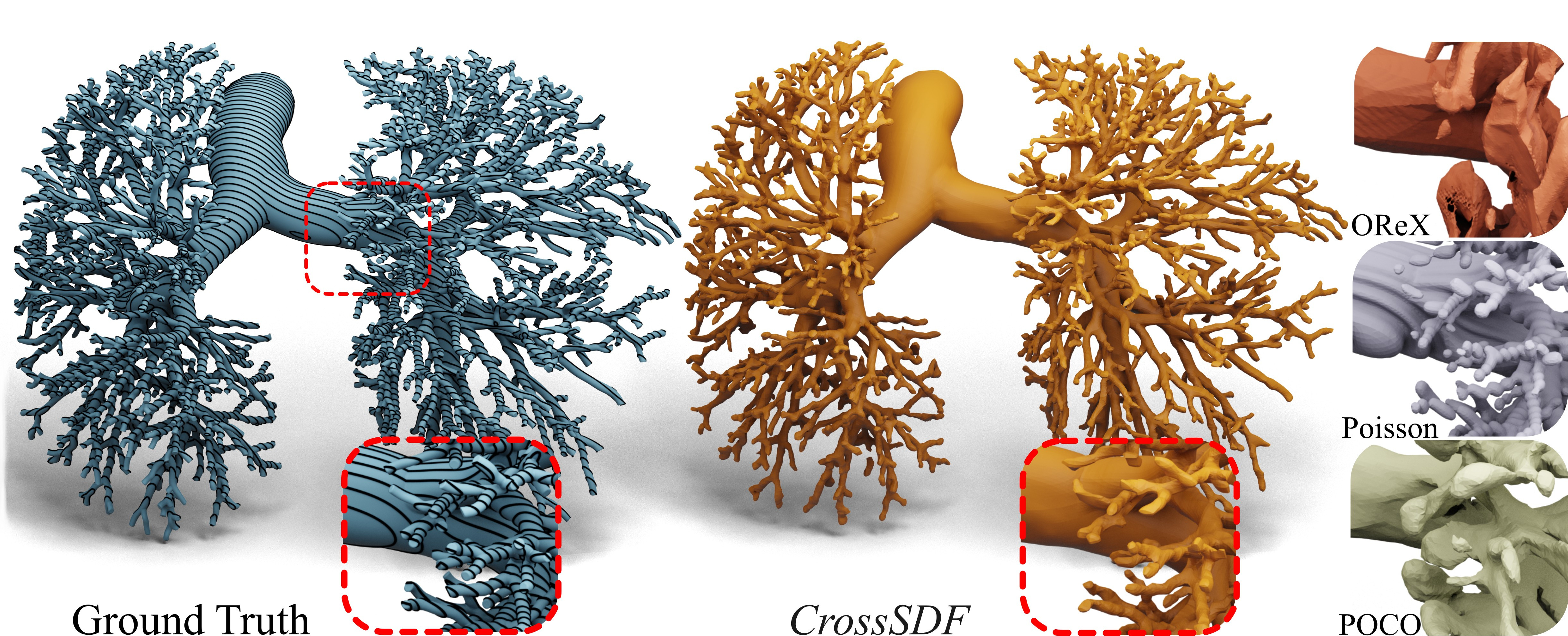} 
\vspace{-6pt}
\captionof{figure}{{}
We propose \textbf{\method},  a novel approach for reconstructing a 3D signed-distance field from 2D cross-sections. 
The input is a set of 2D cross-sections that sample an unobserved ground-truth geometric object by planar intersection (denoted as black lines overlayed on the ground truth \textit{Alveolis} structure on the left). 
\method (middle) accurately reconstructs thin structures without breakages, over-smoothing, or cross-sectional artifacts observed in competing methods (right). 
}%
\label{fig:teaser-fig}%
\end{center}%
}]
\begin{abstract}
 Reconstructing complex structures from planar cross-sections is a challenging problem, with wide-reaching applications in medical imaging, manufacturing, and topography. Out-of-the-box point cloud reconstruction methods can often fail due to the data sparsity between slicing planes, while current bespoke methods struggle to reconstruct thin geometric structures and preserve topological continuity. This is important for medical applications where thin vessel structures are present in CT and MRI scans. This paper introduces \method, a novel approach for extracting a 3D signed distance field from 2D signed distances generated from planar contours. Our approach makes the training of neural SDFs contour-aware by using losses designed for the case where geometry is known within 2D slices. Our results demonstrate a significant improvement over existing methods, effectively reconstructing  thin structures and producing accurate 3D models without the interpolation artifacts or over-smoothing of prior approaches.
\end{abstract}

\section{Introduction}
\renewcommand{\thefootnote}{\fnsymbol{footnote}}
\footnotetext[1]{Equal contribution.}
\footnotetext[2]{Project website: https://iamsalvatore.io/cross\_sdf}
\label{sec:intro}
Reconstructing three-dimensional structures from sparse two-dimensional cross-sectional data is a challenging task often encountered in medical imaging~\cite{liu_surface_2008, jacobs20083d, geiger1993three}, manufacturing~\cite{castellan2000use}, and topography~\cite{TaoJu}. 
In medical imaging in particular, intricate branching structures such as blood vessels and neural tissues must be reconstructed for their correct interpretation in diagnoses and interventions~\cite{3d_blood_vessel}. 
These structures often appear as sparse and incoherent within cross-sectional scan data, which presents a significant problem for reconstruction methods.  
While traditional optimization-based reconstruction techniques designed for slice data can be effective for parallel cross-sections sampled from simple shapes~\cite{bajaj1996arbitrary, barequet1994piecewise, boissonat2007}, they rely on interpolating between cross-sections, which can lead to significant ridging artifacts between the slices~\cite{sawdayee2023orex}. 
Bespoke reconstruction algorithms that handle the more challenging problem of non-parallel cross-sections can scale poorly~\cite{TaoJu, bermano_arbitrary}, limiting their application beyond sparsely-sliced simple geometry. 
The reliance on smoothing algorithms during post-processing also makes them unsuitable for reconstructing thin structures ~\cite{TaoJu}. 
Consequently, there is a need for scalable methods that can reconstruct high-quality geometry, containing thin structures, from arbitrary cross-sections. 

Concurrently, we have seen a significant growth in implicit neural reconstruction methods that characterize shapes as continuous functions of spatial coordinates~\cite{chen2019learning,mescheder2019occupancy,park2019deepsdf}. 
By extracting 3D points from the cross-sections one could frame the task as a point-cloud reconstruction problem. 
However, as seen in \cref{fig:teaser-fig}, current general purpose implicit methods (\eg \cite{wang_neural-imls:_2024,boulch2022poco}) are poorly adapted to the sparsity between slicing planes, leading to gaps, broken geometry, staircase artifacts, and over-smoothing. 
The recently introduced OReX~\cite{sawdayee2023orex} is a neural reconstruction approach specifically designed for cross-sectional data. 
However, due to their indicator field parametrization and spatial sampling pattern, OReX introduces artifacts and misses thin details (see~\cref{fig:teaser-fig}). 

To overcome these limitations, we propose \method, a hash-based~\cite{muller2022instant} neural reconstruction approach that takes a set of 2D cross-sections of an object as input and faithfully reconstructs the full 3D shape. 
We begin by generating 2D signed distance labels from the input cross-sections, which are then used with a novel loss to learn a 3D neural signed distance field (SDF). 
Our loss minimizes the visual artifacts that arise from the sparse nature of the 2D cross-sections, which standard reconstruction methods fail to address.
\method produces high-fidelity 3D reconstructions, excelling in objects containing thin structures. 

Our contributions are as follows:
\begin{itemize}
    \item We introduce a new approach for learning a 3D neural SDF from 2D cross-sections via a novel symmetric difference loss that is capable of representing thin structures without generating unwanted artifacts. 
    \item We propose an adaptive sampling strategy for SDFs points which ensures that thin structures are adequately represented during surface reconstruction.
    \item  We describe a hybrid encoding approach that combines a detail preserving hash encoding with the smoothness of Fourier features to reduce the noticeable grid interpolation artifacts of existing hash-based methods. 
    \item We propose a new challenging benchmark for thin structure reconstruction from cross-sections. 
    \item We obtain state-of-the-art reconstruction results across  a wide variety of both synthetic and real shapes. 
\end{itemize}

\section{Related Work}
\paragraph{3D Reconstruction from Cross-sections.} 
Early methods primarily addressed reconstruction from parallel planes, a scenario often found in medical imaging and topography~\cite{bajaj1996arbitrary, barequet1994piecewise, boissonat2007}. 
These traditional approaches, while effective for simple objects, often rely on explicit interpolation between cross-sections, leading to extreme banding artifacts. 
Later methods were able to accommodate more complex inputs, such as arbitrarily oriented cross-sections, which pose significant challenges in terms of continuity and computational efficiency~\cite{bajaj1996arbitrary, barequet1994piecewise, boissonat2007}. 
Non-parallel cross-sections arise in settings such as freehand 2D ultrasound, where the acoustic beams from the probe produce a set of arbitrary planar cross-sections that can capture finer-scale anatomical structures compared to parallel-only orientations~\cite{shadiamani, barequet2009}. Some optimization-based strategies for arbitrary planes have cubic time complexity in the number of contours involved, limiting their application to sparsely-sliced simple geometry~\cite{bermano_arbitrary}, or they require smoothing post-processing algorithms to produce better results, harming surface accuracy~\cite{TaoJu}. 
Recently, OReX~\cite{sawdayee2023orex} introduced an implicit neural network approach for this problem setting. 
However, their choice of surface representation and positional encoding prevents them from reconstructing fine details and complex structures. 
Our \method approach addresses these limitations, and can handle complex objects to reconstruct surfaces from sparse data with minimal artifacts. 

\begin{figure*}[ht]
\center
\includegraphics[width=0.98\linewidth, trim=0pt 0pt 0pt 0pt, clip]{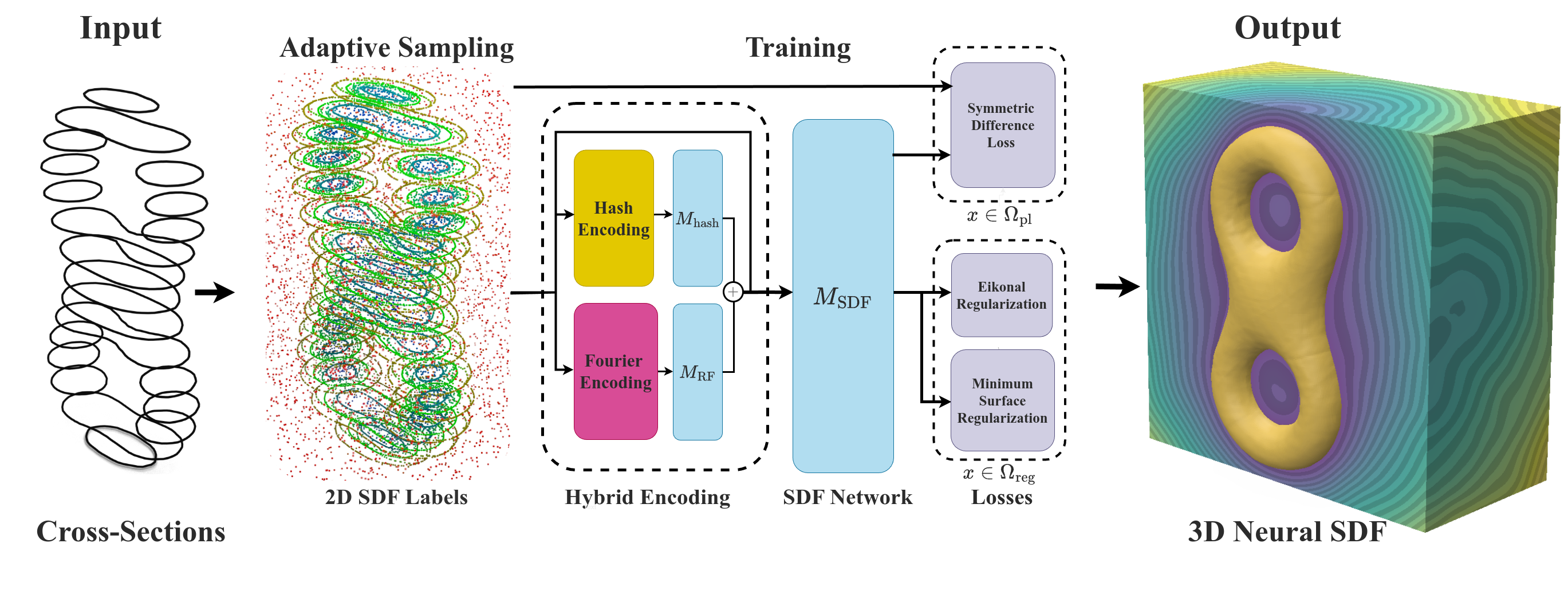} 
\vspace{-25pt}
\caption{\textbf{\method} takes a set of planar cross-sections as input. 
These cross-sections result in contours which denote the  surface boundaries of the target geometry of interest (left, black lines) and each induces a 2D signed distance field (SDF) in its respective plane. 
From this, we generate a set of planar sample points $\Omega_{\text{pl}}$ and their 2D SDF labels. During training, points are encoded using our hybrid encoder before being passed to the SDF network $M_{\text{SDF}}$ for prediction. 
At each iteration we create a set of 3D samples $\Omega_{\text{reg}}$ to apply volumetric regularization. 
The combination of our novel sampling, loss function, and hybrid encoding results in a high quality 3D SDF.
}
\label{fig:overview}
\vspace{-10pt}
\end{figure*}

\paragraph{Neural Surface Reconstruction.} 
Neural networks have found wide applications as a way to smoothly parameterize geometry, offering a compact and  versatile alternative to classical 3D representations. As a result, reconstructing surfaces from raw data is now a well-studied problem, with applications in image-based reconstruction, point cloud reconstruction, and shape-completion~\cite{yariv2023mosaicsdf, xie2022neural, wang2021neus, li2023neuralangelo, boulch2022poco}. 
Of these, reconstructing surfaces from sparse point cloud data is related to our cross-section reconstruction task. 

Due to the difficulty of obtaining surface normals, which are required for classical learning-free methods such as Poisson reconstruction~\cite{kazhdan2013screened, kazhdan2006poisson}, neural network-based approaches have been developed that can operate on unoriented data by predicting the signed distance to a surface~\cite{atzmon2020sal, sitzmann2020implicit}. 
Regularization has also been demonstrated to be effective for reconstructing from sparse data, \eg Eikonal regularization as in~\cite{gropp2020implicit} or implicit-moving-least-squares as in~\cite{wang_neural-imls:_2024}. 
Numerous subsequent improvements have been made by leveraging various properties of SDFs to induce stronger priors~\cite{ma2023towards, yifan2021iso, ma2020neural}. 
Nonetheless, these approaches can struggle with noise as well as maintaining surface continuity when the input is sparse. 

In parallel, data-driven approaches have also been proposed which encode strong priors by training on large datasets to learn the mapping between  point clouds and surfaces~\cite{park2019deepsdf,erler2020points2surf,boulch2022poco}. 
This can result in improved robustness to noise and missing regions. 
Works such as DeepSDF~\cite{park2019deepsdf} optimize a latent space to condition an SDF decoder. Points2Surf~\cite{erler2020points2surf}, and most recently POCO~\cite{boulch2022poco}, directly encode patches through local and global operations, to obtain better generalization. 
However, these methods can still fail on out-of-distribution inputs, \eg complex objects typically captured with cross-sectional data. 

Finally, advancements in image-based reconstruction methods have demonstrated the efficacy of multi-resolution hash-encodings, enabling models to capture fine surface details~\cite{li2023neuralangelo, muller2022instant}. This type of detail is typically lost in sparse point cloud data, but can be present in the dense data found in 2D cross-sections. With this in mind, we leverage hash-encodings to capture the fine details and thin structures captured in the 2D cross-sections of complex objects. 

\section{Method} 

\subsection{Overview}
\paragraph{Input and output.} Our input is a collection $\mathcal{P} = \{ P_1, \dots, P_n \}$ of $n$ 2D planes embedded in $\mathbb{R}^3$ with arbitrary offsets and orientations. On each plane, we have a set of closed contours $\mathcal{C}_i = \{ c_{i}^{1}, \dots, c_{i}^{l_i} \}$. 
The contours are the intersection of the planes and the boundary of the target geometry of interest, and they divide the plane into inside and outside regions. 
Our output is a neural 3D signed distance field (SDF) $f(\mathbf{x}; \boldsymbol{\theta})$, defining the predicted geometry implicitly as its zero set $\mathbf{S} = \{ \mathbf{x} \in \mathbb{R}^3 \mid f(\mathbf{x}; \boldsymbol{\theta}) = 0 \}$. 

\paragraph{Approach.} Previous works like OReX~\cite{sawdayee2023orex} and Bermano \emph{et al.}~\cite{bermano_arbitrary}  define a 2D indicator function on the cross-section planes and interpolate it to obtain a 3D indicator function, where the cross-section operation reproduces the 2D function. However, this is detrimental to the reconstruction task for two main reasons: 1) indicator functions are discontinuous, causing severe normal and smoothness artifacts as shown from OReX qualitative results in \cref{fig:comparisons_thin} and \cref{fig:comparisons_thick}, and 2) training on 2D indicator functions reduces the task to an unstable binary classification problem that easily misclassifies thin features. 
With the objective of our work to make this training stable, smooth, and accurate, we instead learn a 3D SDF, guided by the contours' planar SDFs. The associated Eikonal regularization has been demonstrated to be effective on sparse inputs~\cite{gropp2020implicit}. However, naively fitting a global 3D SDF to the 2D SDFs results in significant artifacts, which we successfully mitigate by only considering the \emph{symmetric difference} between the prediction and the target (Sec.~\ref{sec:symmetric-difference-fitting}). To ensure we capture thin structure, we use an improved sampling strategy (Sec.~\ref{sec:adaptive-sampling}), accompanied by hybrid encoding (Sec.~\ref{sec:hybrid_encoding}) to maintain surface smoothness. 
We summarize our full pipeline in Fig~\ref{fig:overview}.

\subsection{Adaptive Sampling}
\label{sec:adaptive-sampling}
\begin{wrapfigure}{r}{0.33\columnwidth}
    \vspace{-3pt}
    \hspace{-10pt}
    \includegraphics[width=0.33\columnwidth]{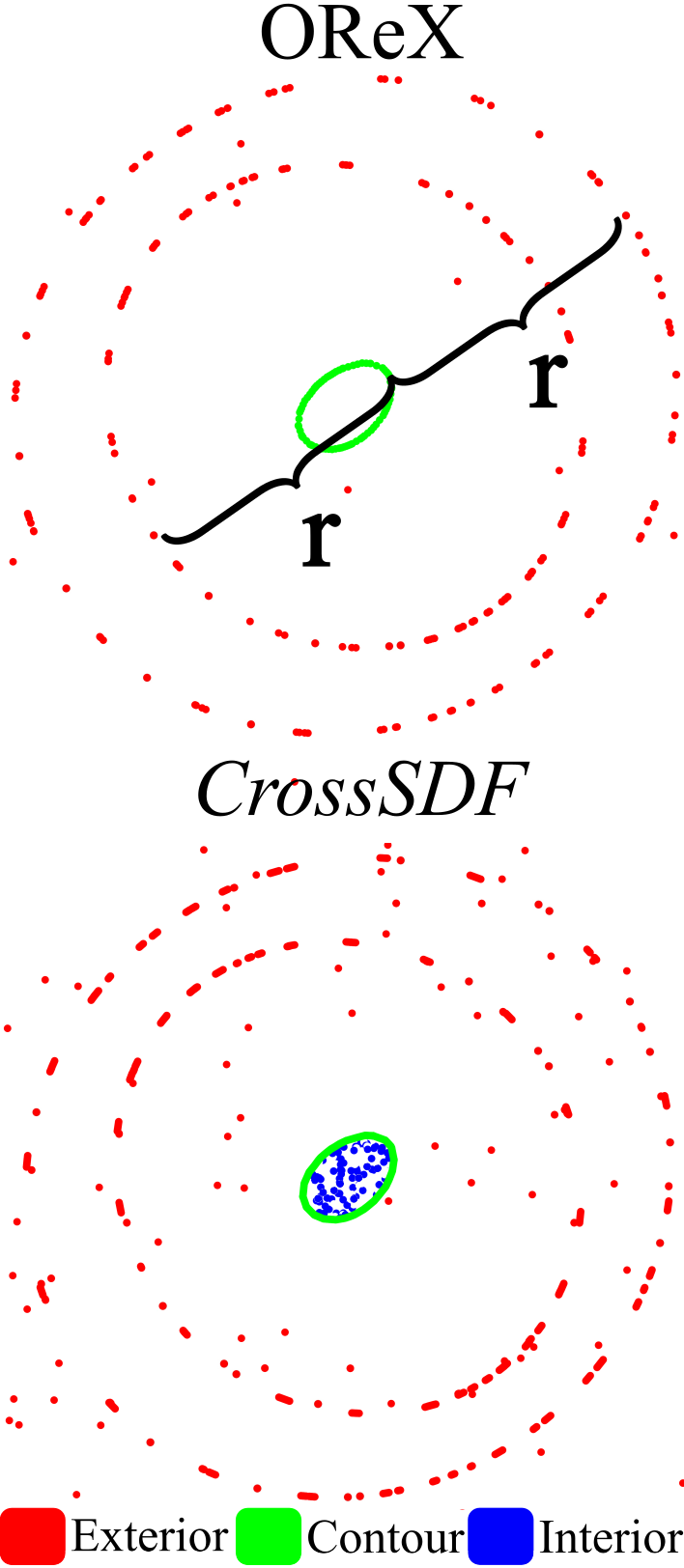}
    \hspace{-10pt}
    \vspace{-11pt}
\end{wrapfigure} 
We define the set of $n$ 2D SDFs as $f_{\text{2D}} = \{ f^i_{\text{2D}}(\mathbf{x}):\mathbb{R}^2 \rightarrow \mathbb{R}\}$,  where $i \in \{1, \dots, n\}$. 
To supervise our training of the 3D SDF $f$, we create a set $\Omega_{\text{pl}}$ of in-plane point samples to 
which we assign the respective (precomputed) $f_\text{2D}$ values. 
We generate in-plane point samples $\Omega_{\text{pl}}$ with thin structures in mind. For these types of objects, where many small contours are present, as seen in~\cref{fig:teaser-fig}, a naive uniform sampling of 2D points on each cross-section would under-sample structures with low cross-sectional area. 
OReX~\cite{sawdayee2023orex} samples points at a fixed radius $r$ at either side of a contour, alongside a uniform sampling of the plane. However, for thin structures, a fixed radius will often extend beyond the contour's interior, leading to disproportionately fewer interior samples (see inset). 
To address this, we adapt the OReX sampling, by directly sampling the interior of every contour until a threshold number of samples is taken. This is made possible by our problem setting which provides ground-truth segmentation of interior and exterior regions on each slicing plane. This per-contour sampling approach ensures the interiors of small contours are adequately sampled, discouraging our model from ignoring contours with small surface area (which we demonstrate in ablations). 

\subsection{Hybrid Encoding}
\label{sec:hybrid_encoding}
Optimizing a neural field directly on 3D coordinates can result in over-smooth geometry, as neural networks are biased towards low-frequency functions on low dimensional domains~\cite{tancik2020fourier}. As a result, positionally encoding 3D input coordinates with Fourier features has become common practice to learn detailed implicit fields~\cite{gao2022nerf}. Grid-based encodings offer an alternative, encoding positions by interpolating learned feature vectors stored on grid vertices \cite{takikawa2021nglod, liu2020neural, li2023neuralangelo}. Recently, the use of spatial hashing has allowed grid-based encodings to scale to higher resolutions, and therefore capture even finer surface details~\cite{muller2022instant,li2023neuralangelo}. 
Inspired by this, we employ a hash-grid encoding to capture thin structures. However, these encodings are not directly suitable to our problem, since they introduce grid artifacts visible in the resulting geometry. These artifacts are due to linear interpolation that is performed across grid cell boundaries, leading to discontinuities that manifest as sharp creases or ridges in the learned surface (see \cref{fig:artifacts_ablation}).
This effect is especially pronounced for our task, since the sparsity of the cross-sections means that large 3D regions between input planes are not intersected and we rely exclusively on priors for interpolation. 
To mitigate these issues, we introduce an alternative encoding strategy combining random Fourier features (RFFs)~\cite{tancik2020fourier} with multi-resolution hash-grid features~\cite{muller2022instant}. 
We define the random RFF encoding \( \gamma_{\text{RF}}: \mathbb{R}^3 \to \mathbb{R}^d \) as follows:
\begin{equation}
\gamma_{\text{RF}}(\mathbf{x}) = [\cos(z_1), \sin(z_1), \dots, \cos(z_{d/2}), \sin(z_{d/2})],    
\end{equation}
where $z_i = \mathbf{b}_i^{\top} \mathbf{x}$ and the coefficients \( \mathbf{b}_i \in \mathbb{R}^3 \) are sampled randomly from a multivariate Gaussian distribution \( \mathcal{N}(\mathbf{0}, \sigma^2 \mathbf{I}) \), with mean \( \mathbf{0} \) and variance \( \sigma^2 \). This sampling ensures a full set of encoding functions spanning a range of frequencies and alignments.

Our hybrid encoding strategy first encodes an input 3D coordinate $\mathbf{x}$ using two different encoding functions; a hash-grid encoding \( \gamma_{\text{hash}}: \mathbb{R}^3 \to \mathbb{R}^d \) and a RFF encoding \( \gamma_{\text{RF}}: \mathbb{R}^3 \to \mathbb{R}^d \). Each encoding is passed through separate one hidden layer MLPs, defined as \( M_{\text{hash}}, M_{\text{RF}}: \mathbb{R}^d \to \mathbb{R}^{d^\prime} \), which are learned, producing intermediate encodings $\mathbf{z}_{\text{hash}} = M_{\text{hash}}( \gamma_{\text{hash}})$ and $\mathbf{z}_{\text{RF}} = M_{\text{RF}}( \gamma_{\text{RF}})$. 
We combine these intermediate encodings by adding them, where the RFF encoding is scaled by a small constant \( \alpha \),
\begin{equation}
    \mathbf{z}_{\text{comb}} = \mathbf{z}_{\text{hash}} + \alpha \cdot \mathbf{z}_{\text{RF}}.
\end{equation}
This constant is a hyperparameter chosen so that the magnitudes of each encoding is approximately equal upon initialization. 
The combined encoding is then concatenated with the original input $\mathbf{x}$ to form the final hybrid encoding input to the SDF MLP, $M_{\text{SDF}}$,
\begin{equation}
\mathbf{z}_{\text{final}} = [\mathbf{z}_{\text{comb}} \mid \mathbf{x}],
\end{equation}
where \( [\cdot \mid \cdot] \) represents concatenation (see~\cref{fig:overview}). The inclusion of random Fourier features encourages smooth representations, reducing interpolation artifacts caused by the discontinuities at hash-grid cell boundaries and improves surface continuity.

\subsection{Optimization}
\label{sec:symmetric-difference-fitting}

\begin{figure}[!tb]
  \centering
  \includegraphics[width=0.85\columnwidth,  trim=0cm 0cm 0cm 0cm, clip]{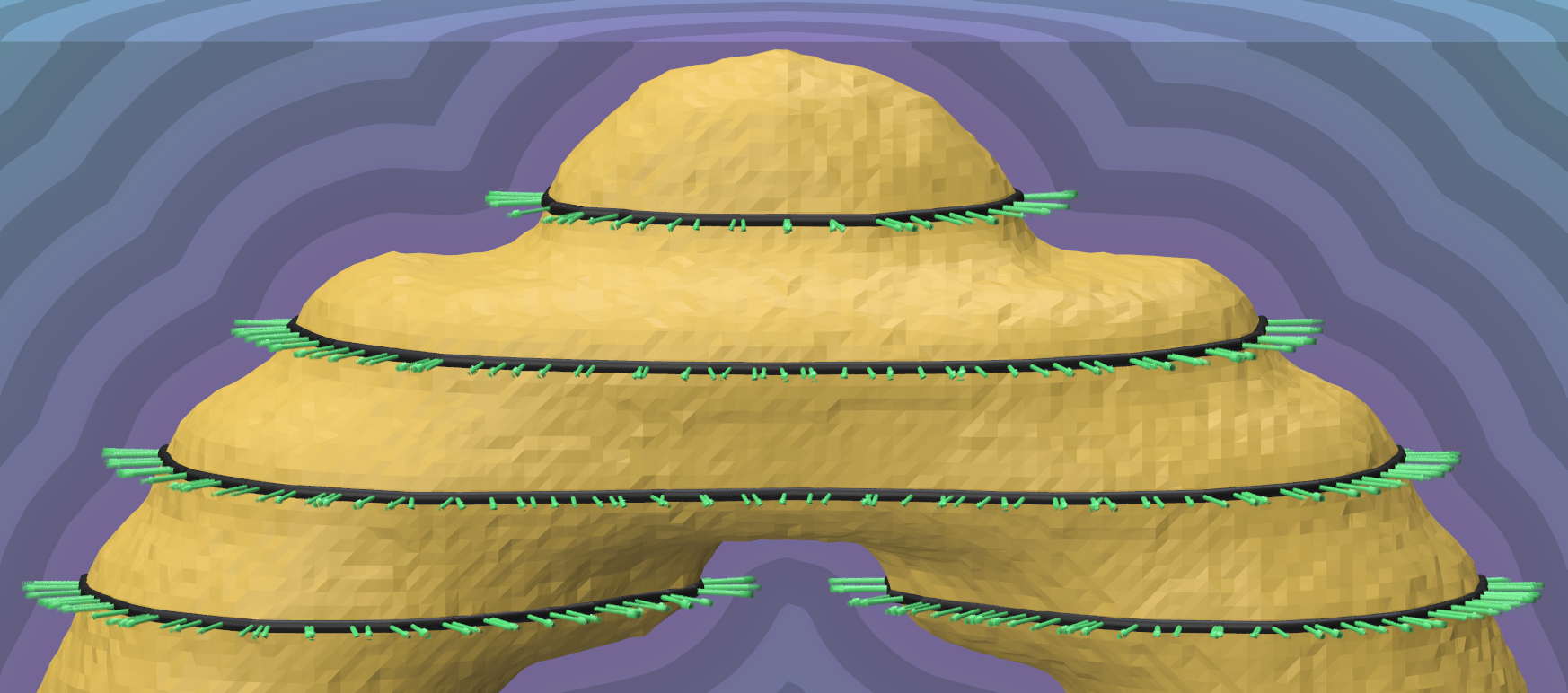}
  \vspace{-7pt}
  \caption{Close-up of the ``figure eight'' object from \cref{fig:overview} when fitting a 3D SDF directly to the \emph{2D SDF} labels. Displayed are the input contours (black) along with 2D SDF gradient vectors (green arrows). Here the neural 3D SDF attempts to remain orthogonal to the cross-sections, resulting in a ``laddering'' effect.}
  \vspace{-12pt}
  \label{fig:signloss}
\end{figure}

\paragraph{Symmetric Difference Loss.}
The 2D SDFs and the desired 3D SDF generally disagree. For parallel slices, fitting the 2D SDF directly will encourage the normals of $f$ and $f_{2D}$ coincide, causing ``laddering'' artifacts (see \cref{fig:signloss} and exhibited in the ablation in~\cref{fig:artifacts_ablation}). For non-parallel slices, there are actual inconsistencies along their intersection in 3D space.
However, the only geometric data each input cross-section contains about the target geometry is the classification of interior/exterior. As such, we do not want the neural field to learn the 2D SDFs directly, but rather use them as a smooth guide to the zero set whilst ultimately learning a 3D SDF. 




To avoid these artifacts, our insight is that we can drive the surface optimization with the 2D SDFs by only considering the region where the target and predicted classifications of interior/exterior disagree. This region shrinks to zero when the target and predicted contours exactly overlap, freeing the neural field from the requirement to be a 2D SDF within the plane. At each iteration, we consider the two disjoint sets,
\begin{align}
     \Omega_{\text{on}} &= \{ \mathbf{x} \in \Omega_{\text{pl}} \mid f_{\text{2D}}(\mathbf{x}) = 0\},\\ \nonumber
    \Omega_{\text{off}} &= \{ \mathbf{x} \in \Omega_{\text{pl}} \mid f_{\text{2D}}(\mathbf{x}) \neq 0, \; \llbracket f(\mathbf{x; \boldsymbol{\theta}}) \rrbracket  \neq \llbracket f_{\text{2D}}(\mathbf{x}) \rrbracket  \},
\end{align}
where $\llbracket \rrbracket $ indicates the sign function.

\begin{wrapfigure}{r}{0.33\columnwidth}
    \vspace{0pt}
    \hspace{-13pt}
    \includegraphics[width=0.33\columnwidth, trim=0.15cm 0cm 0cm 0cm 0cm, clip]{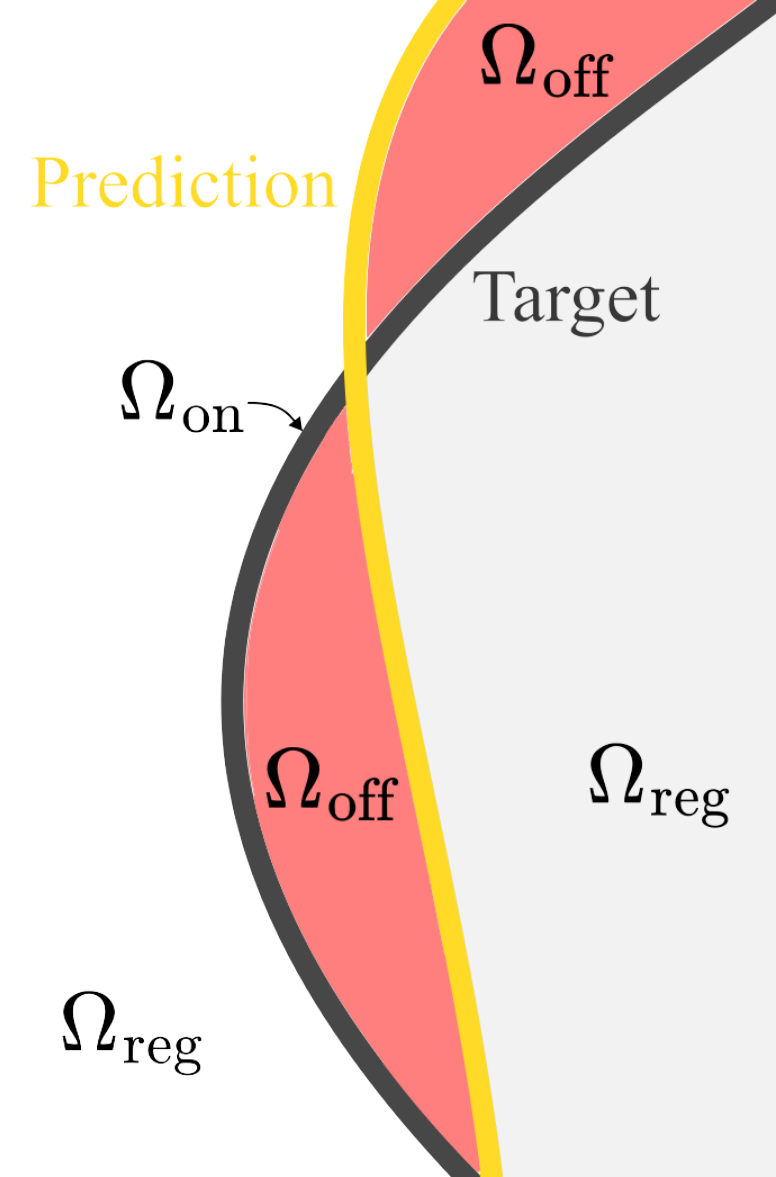}
    \vspace{-10pt}
\end{wrapfigure}
On the right, 
we illustrate where these sample sets are on an example cross-section with target contour (black line) and predicted surface (yellow). The set $\Omega_{\text{on}}$ are on-contour samples. The red regions, where target and predicted classifications of interior/exterior differ, is known as the \textit{symmetric difference} between the two interiors~\cite{symdiff}.   We define separate loss functions on each set,
\begin{align}
\mathcal{L}_{\text{on}} &= \mathbb{E}_{\mathbf{x} \in \Omega_{\text{on}}} \left( \left| f(\mathbf{x}; \boldsymbol{\theta}) - f_{\text{2D}}(\mathbf{x}) \right| \right), \\ \nonumber
\mathcal{L}_{\text{off}} &= \mathbb{E}_{\mathbf{x} \in \Omega_{\text{off}}} \left( (f(\mathbf{x}; \boldsymbol{\theta}) - f_{\text{2D}}(\mathbf{x}))^2\right).
\end{align}
By regressing the 2D SDFs only from samples on the symmetric difference $\Omega_{\text{off}}$ and the contour $\Omega_{\text{on}}$, we can fit the surface while regularizing $f$ be a 3D SDF on the correctly classified regions. This approach successfully mitigates the aforementioned ``laddering'' artifacts and improves reconstruction accuracy. 

\paragraph{Spatial Regularization.} We produce a set $\Omega_{\text{reg}}$ by uniformly sampling points in the full 3D volume. These points are used solely for regularizing $f$, and are generated at each iteration. We use Eikonal regularization to make $f$ a 3D SDF. This regularization term is defined as: 
\begin{equation}
\mathcal{L}_{\text{eik}} = \mathbb{E}_{\mathbf{x} \in \Omega_{\text{reg}}} \left( \|\nabla_{\mathbf{x}} f(\mathbf{x}; \boldsymbol{\theta})\| - 1 \right)^2.
\end{equation}
Additionally, due to the sparse nature of our supervision regions $\Omega_{\text{pl}}$, the model is susceptible to generating spurious surfaces, colloquially known as ``floaters," in regions that are not supervised. To address this issue, we use a minimum surface loss, a regularizer specifically designed to suppress the formation of these unwanted surfaces by penalizing low SDF values~\cite{rebain2021}. We use an exponential term with hyperparameter $\beta$, which penalizes the SDF values that are near zero, and thus discourages the model from generating surfaces at these locations unless strongly supported by the data.
The minimum surface loss is defined as follows:
\begin{equation}
    \mathcal{L}_{\text{min}} = \mathbb{E}_{\mathbf{x} \in \Omega_{\text{reg}}} \left(e^{-\beta|f(\mathbf{x}; \boldsymbol{\theta})|} \right),
\end{equation}
where $f(\mathbf{x})$  represents the predicted SDF value at the point $\mathbf{x}$, and $\Omega_{\text{reg}}$ denotes the regularization domain encompassing the entirety of our input space.  This is important in preventing the emergence of geometrically incorrect features in the reconstructed model, thereby enhancing the SDF predictions.  Our combined training loss is:
\begin{equation}
    \mathcal{L} = \mathcal{L}_{\text{on}} + \mathcal{L}_{\text{off}} + \lambda_{\text{eik}}\mathcal{L}_{\text{eik}} + \lambda_{\text{min}}\mathcal{L}_{\text{min}},
\end{equation}
where $\lambda_{\text{eik}}$ and $\lambda_{\text{min}}$ are regularization weight hyperparameters. 
Please see the Appendix for implementation details.

\begin{figure*}[t]
\center
\includegraphics[width=1\textwidth]{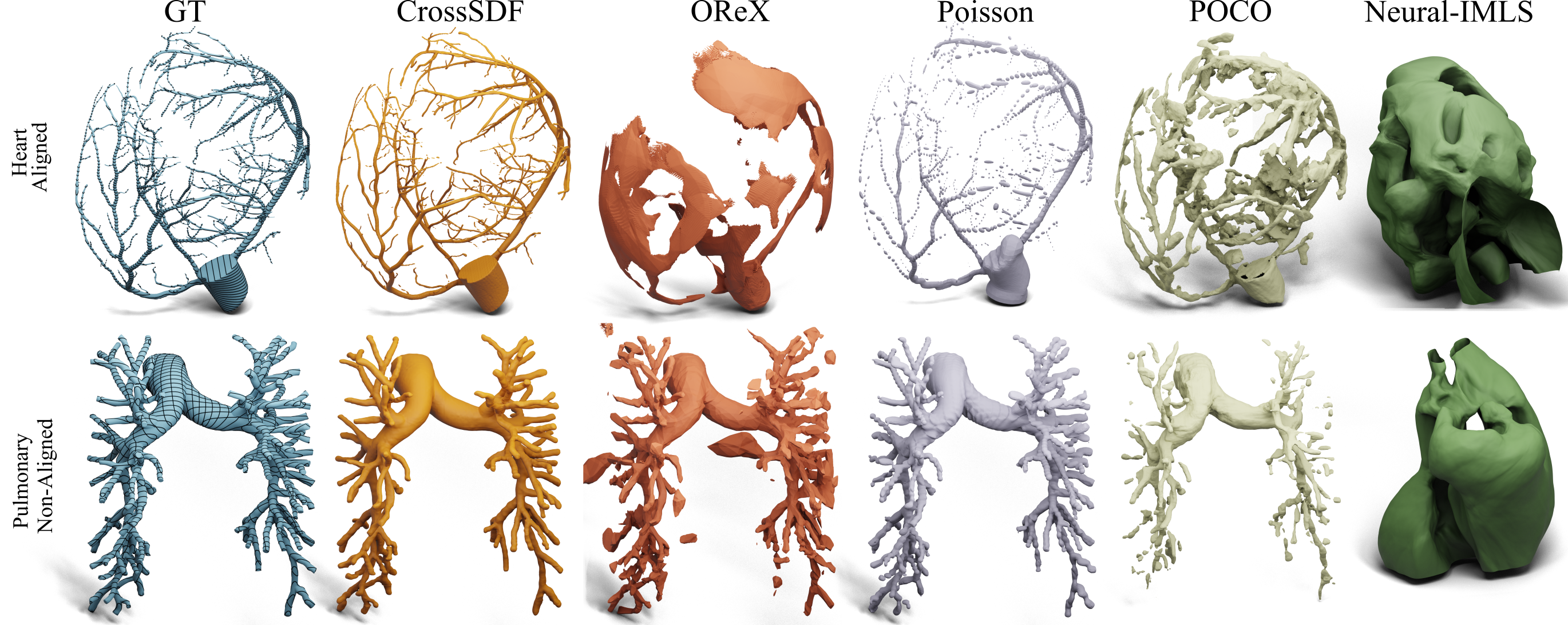}
\vspace{-16pt}
\caption{Reconstruction results on \textbf{thin structures} featuring the Heart (top row) and Pulmonary (bottom row) for various methods. 
Results are presented using both input-aligned and non-aligned planes, displayed on the ground truth meshes.}
\label{fig:comparisons_thin}
\vspace{-8pt}
\end{figure*}

\begin{table*}[t]
    \centering
    \resizebox{\textwidth}{!}{
    {\large
    \begin{tabular}{ll|cccccccccccc}
        \multicolumn{14}{c}{\textbf{Thin Structures}} \\
        \midrule
        & & \multicolumn{2}{c}{Alveolis (100)} & \multicolumn{2}{c}{Cerebral (75)} & \multicolumn{2}{c}{Coronaries (75)} & \multicolumn{2}{c}{Coro (75)} & \multicolumn{2}{c}{Heart (100)} & \multicolumn{2}{c}{Pulmonary (75)} \\
        & Method & Align. & Non-Align. & Align. & Non-Align. & Align. & Non-Align. & Align. & Non-Align. & Align. & Non-Align. & Align. & Non-Align. \\
        \midrule
        \multirow{4}{*}{\rotatebox[origin=c]{90}{\textbf{CD} ($\downarrow$)}} 
        & CrossSDF (Ours) & \textbf{0.35} & \textbf{0.50} & \textbf{0.24} & \textbf{0.26} & \textbf{0.28} & \textbf{0.46} & \textbf{0.21} & \textbf{0.29} & \textbf{0.38} & \textbf{0.82} & \textbf{0.24} & \textbf{0.34} \\
        & OReX~\cite{sawdayee2023orex} & 3.6 & 4.8 & 12 & 2 & 10 & 1.6 & 0.48 & 0.33 & 4.6 & 8.4 & 12 & 5.6 \\
        & Screened Poisson~\cite{kazhdan2013screened} & 0.44 & 0.57 & 0.39 & 0.44 & 0.46 & 0.67 & 1.3 & 0.95 & 1.0 & 0.85 & 0.52 & 0.59 \\
        & POCO~\cite{boulch2022poco} & 4.1 & 4.5 & 0.36 & 0.79 & 1.1 & 0.72 & 1.1 & 3.4 & 2.7 & 4.2 & 2.4 & 2.0 \\
        & Neural-IMLS~\cite{wang_neural-imls:_2024} & 13 & 12 & 14 & 17 & 19 & 17 & 9 & 10 & 19 & 18 & 15 & 15 \\
        \midrule
        \multirow{4}{*}{\rotatebox[origin=c]{90}{\textbf{HD} ($\downarrow$)}}
        & CrossSDF (Ours) & \textbf{14} & \textbf{6.1} & \textbf{3.5} & \textbf{4} & \textbf{5.8} & \textbf{4.7} & \textbf{4.5} & \textbf{2.8} & \textbf{11} & 10 & \textbf{10} & \textbf{7.3} \\
        & OReX~\cite{sawdayee2023orex} & 21 & 35 & 8.1 & 19 & 18 & 27 & 9 & 11 & 26 & 42 & 36 & 21 \\
        & Screened Poisson~\cite{kazhdan2013screened} & 17 & 8.1 & 16 & 25 & 6.1 & 5.4 & 10 & 7.5 & 26 & \textbf{9.2} & \textbf{10} & 7.7 \\
        & POCO~\cite{boulch2022poco} & 32 & 29 & 15 & 41 & 7.1 & 9.2 & 8.8 & 18 & 24 & 25 & 20 & 16.4 \\
        & Neural-IMLS~\cite{wang_neural-imls:_2024} & 68 & 51 & 33 & 48 & 55 & 67 & 37 & 39 & 59 & 61 & 52 & 41 \\
        \midrule
        \multirow{5}{*}{\rotatebox[origin=c]{90}{\textbf{CC}}}
        & \textit{Ground Truth} & 1 & 1 & 1 & 1 & 3 & 3 & 1 & 1 & 1 & 1 & 1 & 1 \\
        & CrossSDF (Ours) & 6 & 33 & 2 & 5 & 5 & 35 & 2 & 4 & 68 & 176 & 2 & 3 \\
        & OReX~\cite{sawdayee2023orex} & 939 & 2119 & 40 & 421 & 219 & 94 & 13 & 4 & 223 & 511 & 914 & 261 \\
        & Screened Poisson~\cite{kazhdan2013screened} & 53 & 1187 & 3 & 23 & 81 & 160 & 9 & 62 & 923 & 1006 & 6 & 17 \\
        & POCO~\cite{boulch2022poco} & 67 & 91 & 2 & 78 & 14 & 46 & 4 & 279 & 70 & 134 & 21 & 134 \\
        & Neural-IMLS~\cite{wang_neural-imls:_2024} & 1 & 1 & 1 & 1 & 1 & 1 & 1 & 1 & 2 & 1 &  \\
        \bottomrule
    \end{tabular}
    }
    }
    \vspace{-8pt}
    \caption{Quantitative results on \textbf{thin structures} across different methods and metrics. 
    The table compares the Chamfer Distance (CD) $\times 100$, Hausdorff Distance (HD) $\times 100$, and number of Connected Components (CC), with the ground truth number also reported, for both aligned and non-aligned versions of each structure. 
    The numbers in parenthesis in the first row denote the numbers of cross-sections. 
    }
    \label{tab:aligned_non_aligned_thin}
    \vspace{-13pt}
\end{table*}

\begin{figure*}[t]
\center
\includegraphics[width=1\textwidth]{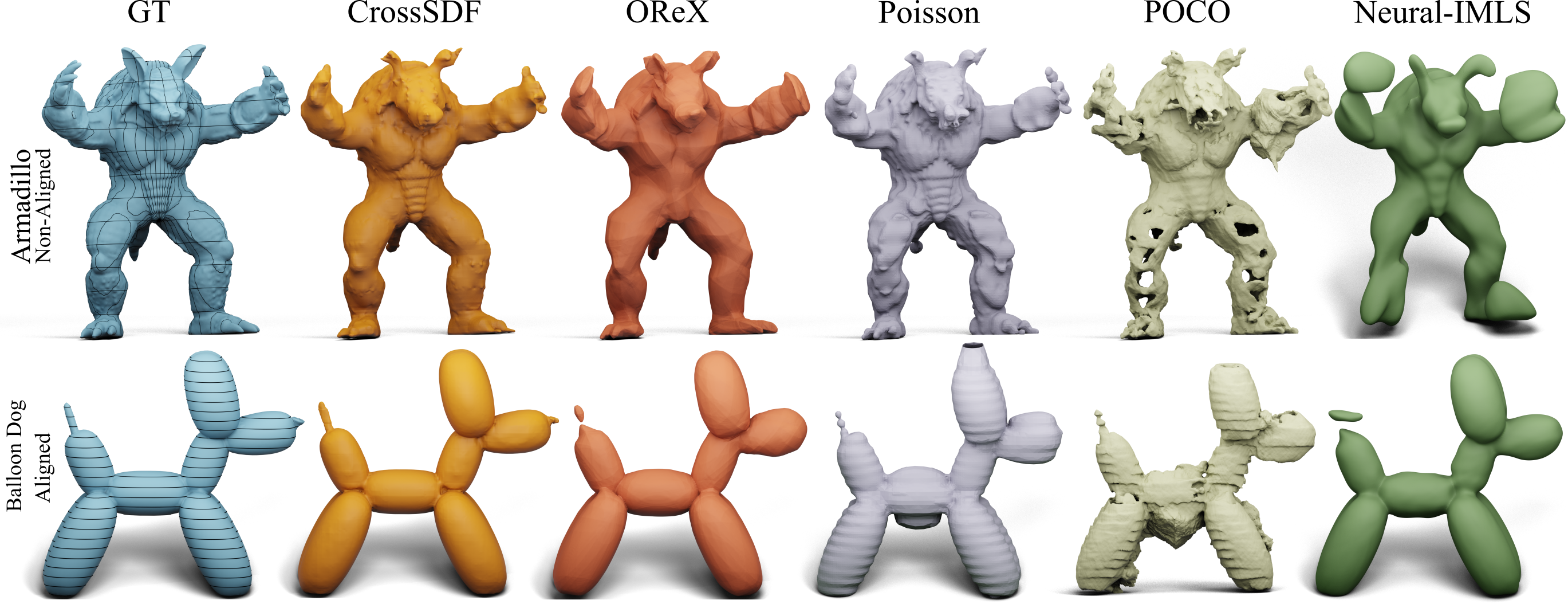}
\vspace{-16pt}
\caption{Reconstruction results on \textbf{thick structures}  featuring the Armadillo (top row) and Balloon Dog (bottom row) for various methods. 
Results are presented using both non-aligned and input-aligned planes, displayed on the ground truth meshes.
}
\label{fig:comparisons_thick}
\vspace{-8pt}
\end{figure*}

\begin{table*}[h]
    \centering
    \resizebox{\textwidth}{!}{
    {\large
    \begin{tabular}{ll|cccccccccccc}
    \multicolumn{14}{c}{\textbf{Thick Structures}} \\
    \midrule
    & & \multicolumn{2}{c}{Armadillo (25)} & \multicolumn{2}{c}{Balloon Dog (25)} & \multicolumn{2}{c}{Elephant (25)} & \multicolumn{2}{c}{Eight (25)} & \multicolumn{2}{c}{Hand (25)} & \multicolumn{2}{c}{Brain (25)} \\
    & Method & Align. & Non-Align. & Align. & Non-Align. & Align. & Non-Align. & Align. & Non-Align. & Align. & Non-Align. & Align. & Non-Align. \\
    \midrule
    \multirow{4}{*}{\rotatebox[origin=c]{90}{\textbf{CD} ($\downarrow$)}} 
    & CrossSDF (Ours) 
    & \textbf{0.80} & \textbf{0.71} & \textbf{0.25} & \textbf{0.23} & \textbf{0.58} & \textbf{0.47} & \textbf{0.10} & \textbf{0.14} & \textbf{0.21} & \textbf{0.27} & \textbf{0.65} & \textbf{0.74} \\
    & OReX~\cite{sawdayee2023orex} 
    & 0.86 & 1.0 & 0.29 & 0.28 & 0.69 & 0.55 & 0.18 & 0.16 & 0.29 & 0.39 & 1.5 & 1.4 \\
    & Screened Poisson~\cite{kazhdan2013screened} 
    & 1.4 & 1.6 & 1.1 & 0.90 & 1.4 & 2.0 &  1.8 & 2.9 & 1.7 & 0.77 & 1.1 & 0.77 \\
    & POCO~\cite{boulch2022poco} & 3.5 & 3.6 & 1.3 & 1.3 & 3.9 & 4.5 & 0.57 & 2.4 & 3.6 & 4.1 & 7.4 & 1.4 \\
    & Neural-IMLS~\cite{wang_neural-imls:_2024} 
    & 1.4 & 3.4 & 0.90 & 0.31 & 0.84 & 0.63 & \textbf{0.10} & 0.15 & 0.35 & 0.46 & 1.4 & 1.6 \\
    \midrule
    \multirow{4}{*}{\rotatebox[origin=c]{90}{\textbf{HD} ($\downarrow$)}}
    & CrossSDF (Ours) 
    & \textbf{7.7} & \textbf{9.0} & \textbf{3.0} & \textbf{3.3} & \textbf{7.1} & \textbf{6.0} & \textbf{1.7} & 1.8 & \textbf{18} & \textbf{16} & \textbf{14} & \textbf{8.5} \\
    & OReX~\cite{sawdayee2023orex} 
    & 8.6 & 9.9 & 6.7 & 4.0 & 7.3 & \textbf{6.0} & \textbf{1.7} & \textbf{1.7} & 22 & 22 & 15 & 14 \\
    & Screened Poisson~\cite{kazhdan2013screened} 
    & 14 & 10 & 8.0 & 6.0 & 8.3 & 6.8 & 13 & 3.9 & 30 & 28 & 19 & 9.1 \\
    & POCO~\cite{boulch2022poco} & 26 & 24 & 14 & 16 & 29 & 29 & 14 & 17 & 21 & 22 & 58 & 15 \\
    & Neural-IMLS~\cite{wang_neural-imls:_2024}  & 16 & 11 & 7.9 & 3.9 & 7.3 & 6.3 & 1.8 & 1.8 & 33 & 29 & 17 & 11 \\
    \bottomrule
    \end{tabular}
    }
    }
    \vspace{-8pt}
    \caption{
    Quantitative results on \textbf{thick structures} across different methods and metrics. 
    The table compares the Chamfer Distance (CD) $\times 100$ and Hausdorff Distance (HD) for both aligned and non-aligned versions of each structure.    
    }
    \label{tab:aligned_non_aligned_thick}
    \vspace{-10pt}
\end{table*}

\section{Experiments} 

\paragraph{Datasets.}
We evaluate \method on both synthetic and real-world datasets. 
We first consider reconstruction from cross-sections  using two synthetic datasets, divided into \textit{thin} and \textit{thick} structures. 
The thin structures dataset consists of six patient-specific,  anatomically accurate, meshes from~\cite{Wilson2013}, originally introduced for  fluid and solid mechanics evaluation which we adapt for 3D reconstruction. 
It contains complex, thin branching geometry, which is particularly challenging for reconstruction methods. 
We generate 75--100 cross-sections for each, which despite being relative dense, still presents a significant challenge. 
The thick structures dataset consists of six meshes used in OReX~\cite{sawdayee2023orex} which have simple topologies, large smooth geometric features, and result in contours with large cross-sectional areas. 
We take sparse cross-sections (\ie 25) to produce the input contours, testing the models' ability to smoothly interpolate sparse inputs without producing plane-slicing artifacts or unwanted gaps. 
For both datasets, we also consider \textit{aligned} and \textit{non-aligned} setups to test the models' ability to handle arbitrary cross-section orientations. For the aligned case, the cross-sections are a set of axis-aligned parallel planes. For the non-aligned case, the planes consist of half parallel and half rotated, around an axis, planes. 

We also test on real word data from the IRCADb-01~\cite{soler20103d} and medical decathlon~\cite{Antonelli_2022} datasets, which contains anonymized and annotated CT scans in DICOM format. As no ground truth 3D geometry is available, we created training and test splits by withholding $\approx 10\%$ of the slices.

\paragraph{Baselines.}
We compare to the state-of-the-art (SoTA) non-data-driven point cloud reconstruction method Neural-IMLS~\cite{wang_neural-imls:_2024}, SoTA data-driven point cloud reconstruction method POCO~\cite{boulch2022poco}, the bespoke cross-section reconstruction approach OReX~\cite{sawdayee2023orex}, and classical Screened Poisson Reconstruction~\cite{kazhdan2013screened}. 
Since screened Poisson reconstruction requires oriented points, we provide it with additional vertex normals which are computed for each point as the average of the connecting edge normals within the cross-sectional plane. 
Implementation details for \method can be found in the Appendix. 

\paragraph{Evaluation.}
We report the Chamfer Distance (CD) and Hausdorff Distance (HD) compared to the ground truth mesh for all structures, and also the number of Connected Components (CC) for the thin structures. 
HD assesses the maximum discrepancy between two 3D meshes, which is important for precision-critical applications like medical imaging. 
CD complements this by averaging nearest point distances to gauge overall shape similarity. 
Evaluating the number of CC confirms correct segmentation and topological consistency across meshes which is important for maintaining structural integrity. 
Together, these metrics provide a thorough framework for precise 3D mesh analysis. 
While we refer the reader to \cite{oechsle2021unisurf} for definitions. 
However,  no single metric fully captures the perceptual quality of the reconstructions. 
For example,  CD gives a good sense of overall similarity, it smoothens out small deviations produced by surface details or ridging artifacts. 


\subsection{Results} 
\cref{tab:aligned_non_aligned_thin,tab:aligned_non_aligned_thick} contain quantitative results for both thick and thin structures. 
Qualitative results, for a subset of structures, can be found in \cref{fig:comparisons_thin,fig:comparisons_thick}. 
The data-driven POCO performs poorly, particularly on thick structures where the sparsity between planes leads to unwanted gaps (\eg Armadillo in~\cref{fig:comparisons_thick}), likely due to being an out-of-distribution input. 
Neural-IMLS  performs reasonably well on thick structures but suffers from over-smoothing, failing to converge when trained on the thin medical structures. 
Note, this over-smoothing manifests as a single blob, which results in artificially good performance on the CC metric. 
OReX also fails to resolve these structures likely as a result of its indicator function representation, sampling issues, and choice of positional encoding. 
Screened Poisson exhibits significant artifacts in the form of ridging and growths, especially on thick structures. 
Quantitatively, it performs well on the denser sliced thin structures, particularly on Hausdorff distance. 
This is likely because this metric is very sensitive to outliers, and Screened Poisson does not produce spurious ``floaters" which can be seen from implicit neural methods on challenging geometry. 
Despite this, it results in a highly fragmented reconstructions that are topologically far from the ground truth. 
It is also worth highlighting that its ridging artifacts  are potentially  harmful in the context of medical interpretation, despite not being very evident from the quantitative scores (\eg Pulmonary in \cref{fig:comparisons_thin}).

\begin{figure}[t!]
\centering
\includegraphics[width=1\columnwidth, trim=0cm 0cm 1.4cm 0cm, clip]{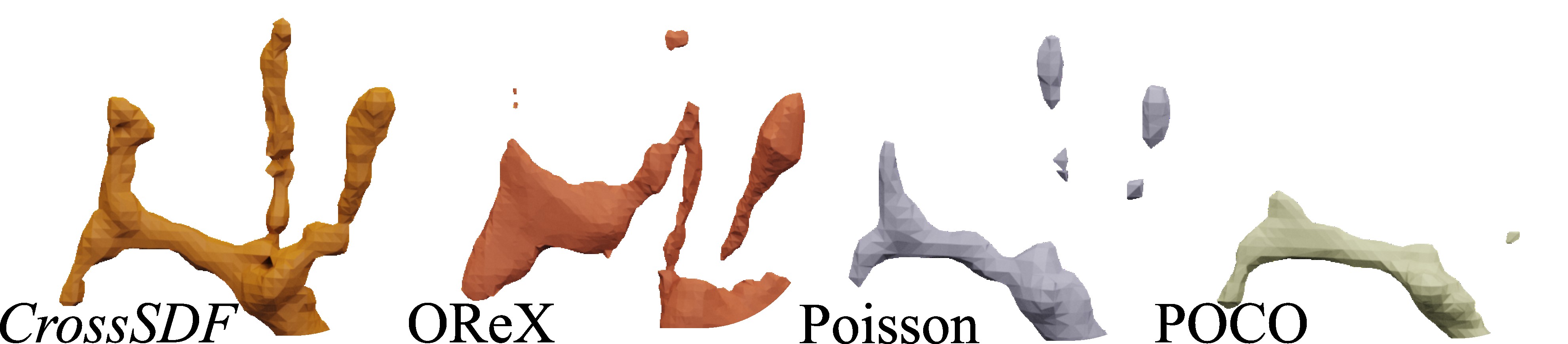}   
\vspace{-14pt}
   \caption{Qualitative comparison of hepatic vessel reconstruction (partial view shown), with 61 slices derived from a real CT scan.}
   \vspace{-20pt}
\label{fig:real_example}
\end{figure}

\begin{table}[h]
    \centering
    \resizebox{\columnwidth}{!}{
    \begin{tabular}{l|c|c|c|c}
        \toprule
        \multirow{3}{*}{\textbf{Method}} & \multicolumn{2}{c|}{\textbf{Hepatic Vessel}} & \multicolumn{2}{c}{\textbf{Liver}} \\
        & Patient 31 & Patient 33 & Patient 13 & Patient 14 \\
        & (61 Slices) & (71 Slices) & (149 Slices) & (75 Slices) \\
        
        \midrule
        CrossSDF (Ours) & \textbf{0.79} & \textbf{0.63} & \textbf{0.83} & \textbf{0.88} \\
        OReX \cite{sawdayee2023orex} & 0.71 & 0.09 & 0.77 & 0.81 \\
        Poisson \cite{kazhdan2013screened} & 0.24 & 0.55 & 0.21 & 0.84 \\
        \bottomrule
    \end{tabular}
    }
    \vspace{-5pt}
    \caption{Results on real CT scans. We report the average 2D IoU on a held out set of evaluation slices.}
    \label{tab:medical_results}
    \vspace{-8pt}
\end{table}

In almost all cases, \method significantly outperforms competing baselines across all evaluation metrics and cross-section types. 
It generates accurate surfaces for the thick structures while  still reconstructing  details which are smoothed over by competing methods, \eg the stomach region of the Armadillo in \cref{fig:comparisons_thick}. 
It also does not produce ridging artifacts unlike many of the baselines (\eg  Screened Poisson) in the context of  sparse cross-sections, \eg the Balloon Dog in \cref{fig:comparisons_thick}. 
On thin structures, \method reconstructs detailed geometry and connected branches where other methods fail. 

\paragraph{CT Scans.} \method displays superior performance on real-world CT scan data, see qualitative example in \cref{fig:real_example}, and quantitative results in \cref{tab:medical_results}. Please see the Appendix for more results.

\vspace{0pt}
\begin{figure}[t]
\includegraphics[width=1\columnwidth, trim=0cm 0cm 0cm 0cm, clip]{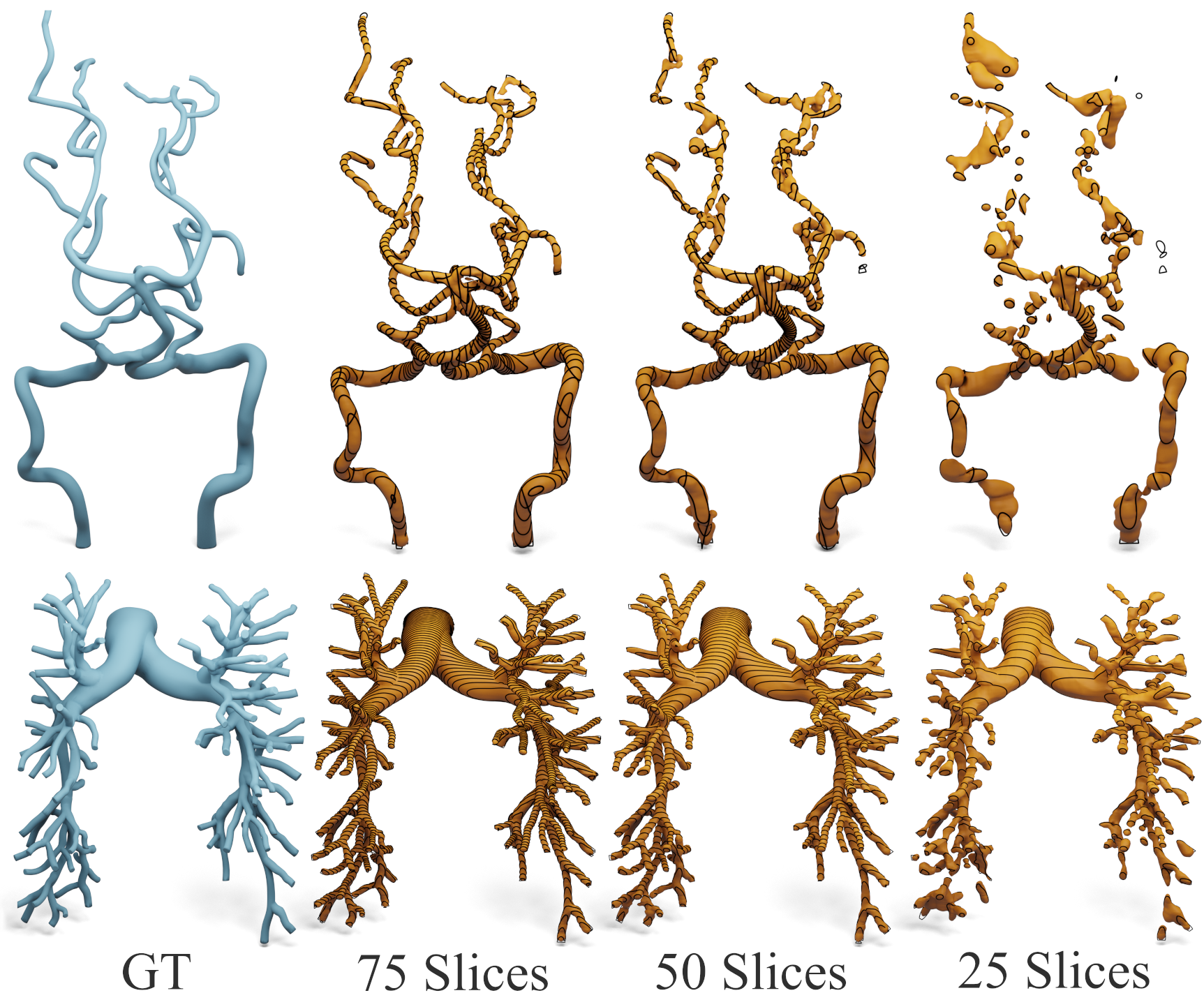}   
\vspace{-18pt}
   \caption{Impact of reducing the number of cross-sections on the Cerebral (top) and Pulmonary (bottom) structures using  non-aligned and aligned cross-sections, respectively. The input cross-sections are denoted in black on the reconstructions (orange).}
   \vspace{-4pt}
\label{fig:slices_ablation}
\end{figure}

\begin{figure}[!htb]
\centering
\includegraphics[width=0.93\columnwidth]{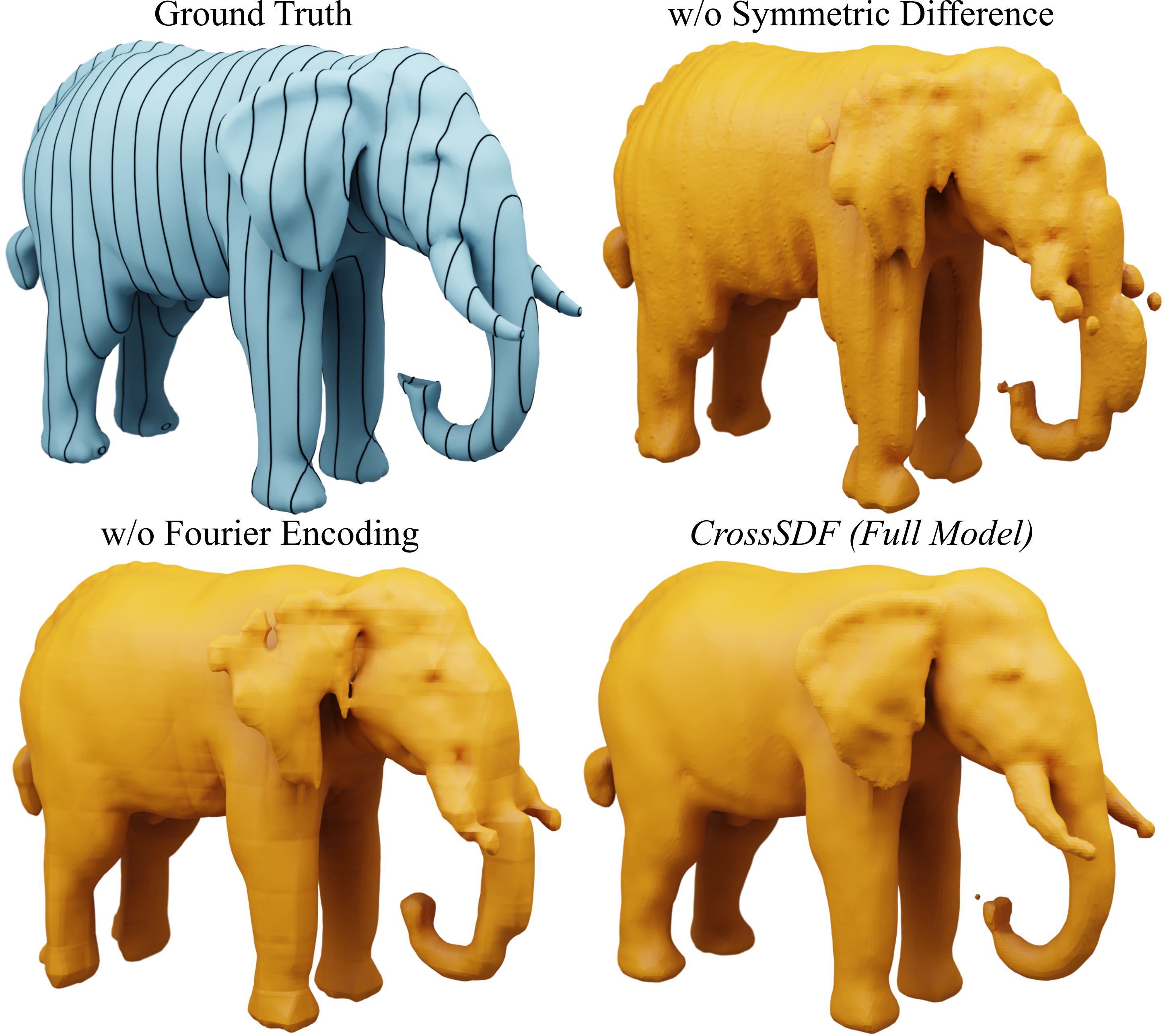}
\vspace{-10pt}
\caption{Qualitative ablation of \method with different components removed. 
The full model results in fewest visual artifacts. 
}
\label{fig:artifacts_ablation}
\vspace{-14pt}
\end{figure}

\subsection{Ablations}
We perform the following ablations to evaluate the impact of  removing various core components from our method:
\begin{itemize} 
    \item \textbf{w/o Fourier Encoder:} Instead of incorporating Fourier features via our hybrid encoding method outlined in~\cref{sec:hybrid_encoding}, we use a vanilla hash-encoding. 
    \item \textbf{w/o Symmetric Difference:} Instead of our  symmetric difference loss from \cref{sec:symmetric-difference-fitting}, we use an L1 loss between the predicted and ground truth 2D SDF labels. 
    \item \textbf{w/o Adaptive Sampler:} Instead of the adaptive sampling introduced in~\ref{sec:adaptive-sampling}), we use the sampling from OReX~\cite{sawdayee2023orex}.
\end{itemize}

\begin{figure}[h]
\centering
\includegraphics[width=0.9\columnwidth, trim=0cm 0cm 0cm 0cm, clip]{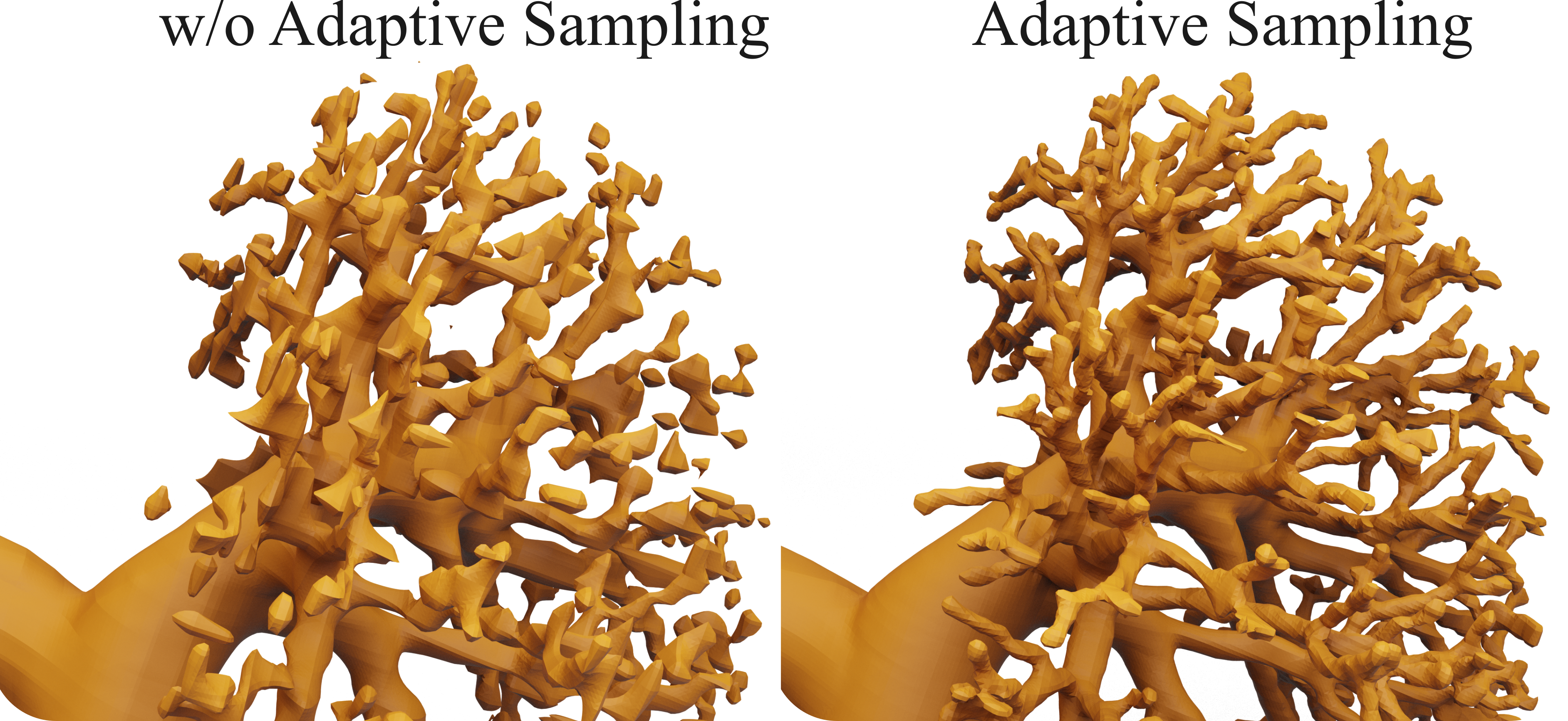}   
\vspace{-8pt}
   \caption{Ablation of our adaptive sampling approach for Alveolis reconstruction.}
   \vspace{-6pt}
\label{fig:real_ablation}
\end{figure}

We present the quantitative results in \cref{tab:artifact_ablation_combined} and visualize them in 
\cref{fig:artifacts_ablation}. Removing our hybrid encoding results in grid interpolation artifacts and loss of surface continuity (see elephant ear and stomach). When our Symmetric Difference loss is removed we observe strong staircase artifacts, which is particularly harmful on the thick structures. For the thin structures, the adaptive sampler is very important for reconstructing thin details (see \cref{fig:real_ablation}). 

Additionally, we also evaluate \method's robustness when the number of input cross-sections are reduced. 
Quantitative and qualitative results can be found in \cref{tab:slices_ablation_results} and \cref{fig:slices_ablation}.  
Unsurprisingly, performance decreases when the number of slices are reduced, but the results are not too different from  the ground truth for $\geq50$ slices, and \method outperforms OReX in all cases.

\begin{table}[t]
    \centering
    \resizebox{\columnwidth}{!}{
    \begin{tabular}{l|cc|cc|cc|cc}
        \toprule
        & \multicolumn{2}{c|}{\textbf{Balloon Dog}} & \multicolumn{2}{c|}{\textbf{Elephant}} &  \multicolumn{2}{c|}{\textbf{Alveolis}} &  \multicolumn{2}{c}{\textbf{Coronaries}} \\
        \textbf{Model} & \textbf{CD} ($\downarrow$) & \textbf{HD} ($\downarrow$) & \textbf{CD} ($\downarrow$) & \textbf{HD} ($\downarrow$) & \textbf{CD} ($\downarrow$) & \textbf{HD} ($\downarrow$) & \textbf{CD} ($\downarrow$) & \textbf{HD} ($\downarrow$) \\
        \midrule
         \textit{CrossSDF (full model)} & \textbf{0.24} & \textbf{3.0} & 0.58 & \textbf{7.1} & \textbf{0.35} & \textbf{14} & \textbf{0.28} & \textbf{5.8}\\
         w/o Fourier Encoding & 0.30 & 3.4 & 0.88 & 10.5 & 0.41 & 15 & 0.33 & 6.6\\
         w/o Symmetric Diff. & 0.57 & 4.1 & 1.01 & 14.4 & 0.38 & 15 & 0.29 & 6.1\\
         w/o Adaptive Sampler & \textbf{0.24} & 3.1 & 0.58 & \textbf{7.1} & 0.82 & 16 & 0.55 & 12.5 \\
         w/o $\mathcal{L}_{min}$ & 0.31 & 6.1 & 0.77 & 13 & 0.54 & 18 & 0.49 & 14 \\
         only Fourier & 0.27 & 3.1 & \textbf{0.57} & 7.4 & 0.94 & 17 & 0.45 & 11 \\
         L1 on $\Omega_\text{off}$ & 0.25 & 4.3 & 0.58 & 7.3 & \textbf{0.35} & \textbf{14} & \textbf{0.28} & 6.2 \\
        \bottomrule
    \end{tabular}
    }
    \vspace{-6pt}
    \caption{Ablation of \method with different components removed, on thick (Balloon Dog and Elephant)  and thin (Alveolis and Coronaries) structures using aligned planes. 
    } 
    \label{tab:artifact_ablation_combined}
    \vspace{-4pt}
\end{table}

\begin{table}[t]
    \centering
    \resizebox{\columnwidth}{!}{
    \begin{tabular}{l|l|cc|cc|cc}
        \toprule
        \textbf{Scene} & \textbf{Method} & \multicolumn{2}{c}{\textbf{75 Slices}} & \multicolumn{2}{c}{\textbf{50 Slices}} & \multicolumn{2}{c}{\textbf{25 Slices}} \\
        & & \textbf{CD} ($\downarrow$) & \textbf{HD} ($\downarrow$) & \textbf{CD} ($\downarrow$) & \textbf{HD} ($\downarrow$) & \textbf{CD} ($\downarrow$) & \textbf{HD} ($\downarrow$) \\
        \midrule
        \multirow{2}{*}{Cerebral} & CrossSDF & \textbf{0.26} & \textbf{4} & \textbf{0.51} & \textbf{16} & \textbf{2.1} & \textbf{26} \\
                                  & OReX & 2 & 19 & 2.6 & 18 & 5.9 & 72 \\
        \midrule
        \multirow{2}{*}{Pulmonary} & CrossSDF & \textbf{0.24} & \textbf{10} & \textbf{0.33} & \textbf{11} & \textbf{1.0} & \textbf{13} \\
                                   & OReX     & 12 & 36 & 14 & 69 & 18 & 78 \\
        \bottomrule
    \end{tabular}
    }
    \vspace{-5pt}
    \caption{Ablation reducing the number of input cross-section for our \method approach and OReX .}
    \label{tab:slices_ablation_results}
    \vspace{-10pt}
\end{table}

\paragraph{Limitations.}
Even though \method successfully reconstructs  challenging structures, it exhibits some limitations. 
Like the other baselines, it will still fail to accurately reconstruct an object given only a very sparse set of slices. 
It could benefit from data-driven priors to improve accuracy. 
In addition, the predictions from such a model would be faster compared to our optimization-based approach.

\section{Conclusion}
We introduced \method, a novel approach for reconstructing detailed 3D structures from a set of arbitrary 2D planar cross-section inputs. 
It relies on several contributions that are specifically targeted towards reconstructing thin structures, including  a novel symmetric difference loss, adaptive contour sampling, and a hybrid feature encoding. 
Combined, these components result in superior qualitative reconstructions, specifically thin structures, and  significantly reduced visual artifacts. 
We introduced a new benchmark for quantitatively evaluating thin structure reconstruction and presented results on both real and synthetic data and from aligned and non-aligned planes. 

\clearpage
\noindent\textbf{Acknowledgments.} SE was supported by the UKRI CDT in Biomedical AI.

{
    \small
    \bibliographystyle{ieeenat_fullname}
    \bibliography{main}
}

\clearpage
\appendix 


\maketitlesupplementary
\setcounter{page}{1}

\setcounter{table}{0}
\renewcommand{\thetable}{A\arabic{table}}
\setcounter{figure}{0}
\renewcommand{\thefigure}{A\arabic{figure}}

\section{Additional Results}

\subsection{CT Scan Results}

\begin{figure*}[t]
\centering
\includegraphics[width=1\textwidth, trim=5pt 0pt 30pt 0pt, clip]{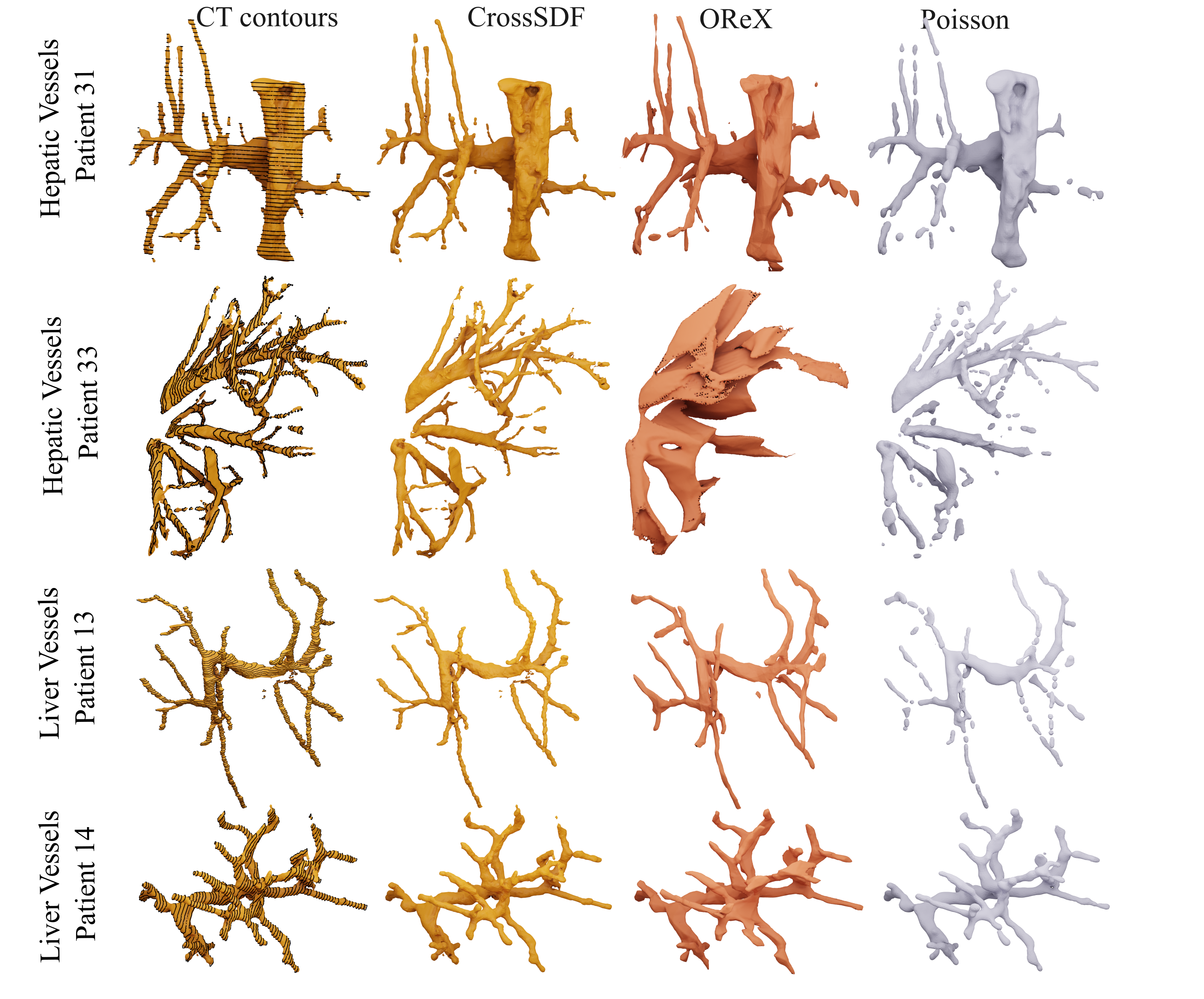}
\vspace{-25pt}
\caption{Additional qualitative results on real CT vessel structures. We compare our \method approach with existing methods. Note the `CT contours' are overlayed on our result to improve viewing clarity.
}
\label{fig:supp_ct}
\end{figure*}

We present qualitative reconstructions of real anatomical structures with two vessel structures from the IRCADb-01 dataset \cite{soler20103d} and two from the Medical Decathlon dataset \cite{Antonelli_2022}, using pixel-level human-annotated segmentations. We evaluate \method on hepatic vessels from slices spaced 5mm apart, and liver vessels at 1mm apart. We display the results across different methods in \cref{fig:supp_ct} which  demonstrates our model's ability to handle clinically relevant challenges. 
As no ground truth 3D geometry is available, we created training and test splits by withholding $\approx 10\%$ of the slices and reported the 2D intersection-over-union (IoU) on the test set. To do so, we use integer division to compute a slice skipping frequency to withhold cross-sections. For example, for the hepatic vessel of patient 31, we remove every $61 // 10 = 6^{th}$ cross-section to compute 2D IoU. 
The quantitative results can be found  in \cref{tab:medical_results}. 

Similarly to the synthetic scenes present in the main paper, OReX~\cite{sawdayee2023orex} generally results in overly smooth geometry. 
In some cases, such as the thin and highly-branched contours of Patient 33's Hepatic Vessel, OReX fails to converge to a good reconstruction. 
Screened Poisson~\cite{kazhdan2013screened} handles these cases more faithfully but is prone to breakages from the skipped slices, which is evident both qualitatively and quantitatively. Finally, we note CT scans with sparse slice thicknesses  are particularly difficult to reconstruct, highlighting our model's robustness in such scenarios~\cite{ahmedm_2024}.

        

\begin{table*}[t]
    \centering
    \resizebox{\textwidth}{!}{%
    \begin{tabular}{ll|cccccccccccc}
        \multicolumn{14}{c}{\textbf{Thin Structures}} \\
        \midrule
        & & \multicolumn{2}{c}{Alveolis (100)} & \multicolumn{2}{c}{Cerebral (75)} & \multicolumn{2}{c}{Coronaries (75)} & \multicolumn{2}{c}{Coro (75)} & \multicolumn{2}{c}{Heart (100)} & \multicolumn{2}{c}{Pulmonary (75)} \\
        & Method & Align. & Non-Align. & Align. & Non-Align. & Align. & Non-Align. & Align. & Non-Align. & Align. & Non-Align. & Align. & Non-Align. \\
        \midrule
        \multirow{5}{*}{
          \rotatebox[origin=c]{90}{
            \begin{tabular}{c}
              \rotatebox[origin=c]{-90}{\textbf{IoU}}\\
                \rotatebox[origin=c]{-90}{$\uparrow$}\\
                \end{tabular}
          }
        } 
        & Neural-IMLS~\cite{wang_neural-imls:_2024} & 0.050 & 0.048 & 0.44 & 0.045 & 0.029 & 0.040 & 0.71 & 0.74 & 0.033 & 0.021 & 0.056 & 0.045 \\
        & Screened Poisson~\cite{kazhdan2013screened} & 0.84 & 0.72 & 0.85 & 0.79 & 0.78 & 0.68 & 0.85 & 0.89 & 0.28 & \textbf{0.71} & 0.89 & 0.86 \\
        & OReX~\cite{sawdayee2023orex} & 0.37 & 0.12 & 0.64 & 0.33 & 0.51 & 0.63 & 0.96 & \textbf{0.97} & 0.32 & 0.42 & 0.59 & 0.74 \\
        & CrossSDF (Ours) & \textbf{0.87} & \textbf{0.76} & \textbf{0.94} & \textbf{0.90} & \textbf{0.88} & \textbf{0.78} & \textbf{0.99} & \textbf{0.97} & \textbf{0.77} & 0.70 & \textbf{0.96} & \textbf{0.94} \\
        \bottomrule
    \end{tabular}%
    }
    \caption{Volume IoU comparison across thin structures.}
    \label{tab:volume_iou_thin}
\end{table*}

\subsection{Additional Synthetic  Results}
\begin{figure*}[t]
\centering
\includegraphics[width=1\textwidth]{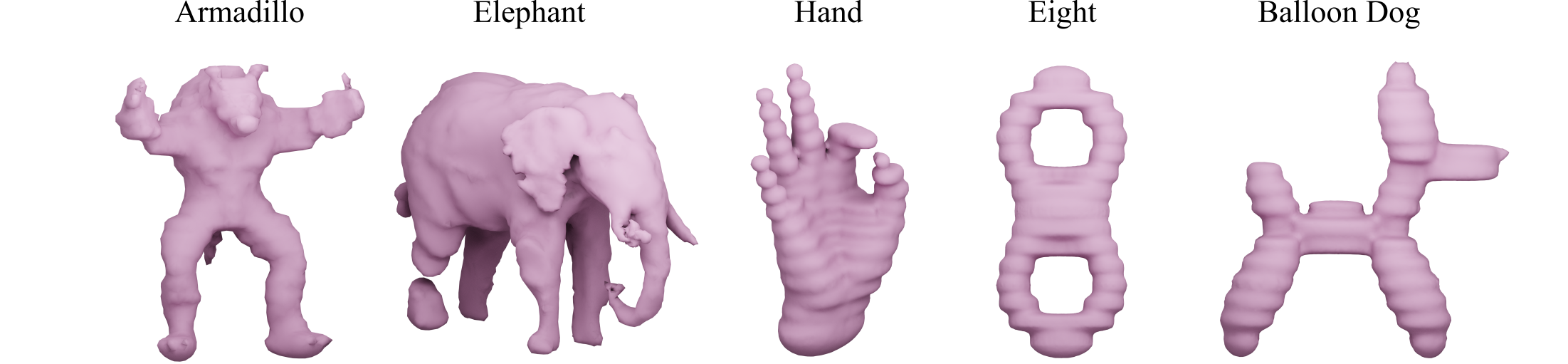}
\vspace{-14pt}
\caption{Additional qualitative results from the method outlined in Bermano \etal \cite{bermano_arbitrary} using our synthetic dataset. Note the extreme laddering artifacts.
} 
\label{fig:bermano_supp}
\vspace{-10pt}
\end{figure*}

\paragraph{Additional Comparisons.} 
We display further qualitative comparisons in \cref{fig:supp_thin,fig:supp_thick}. We also report quantitative comparisons of 3D Volume IoU in \cref{tab:volume_iou_thin}. In \cref{fig:bermano_supp} we also show results from Bermano \etal \cite{bermano_arbitrary}, a non-neural method that can handle arbitrary cross-section orientations, and works by a barycentric blending of indicator functions in the cells of the planar arrangements of the cross-sections. As can be seen in the figure, their  approach is susceptible to laddering artifacts, and fails to run on more complex scenes due to poor scalability.

\begin{figure*}[htbp]  
\centering
\includegraphics[width=1\textwidth]{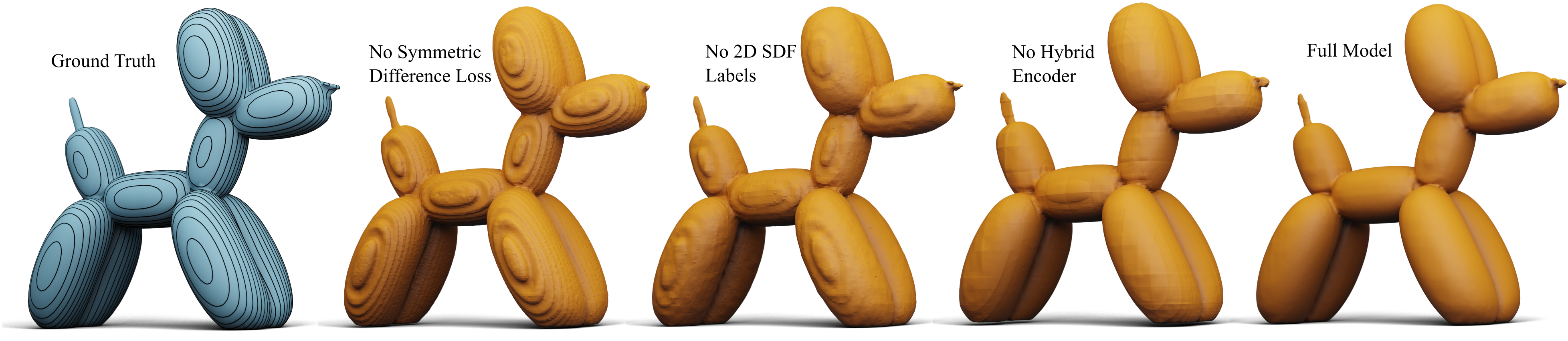}
\vspace{-20pt}  
\caption{Additional qualitative ablation results on the Balloon Dog scene. 
} 
\label{fig:ablation_dog}
\vspace{5pt}  
\end{figure*}

\paragraph{No 2D SDF Labels.} 
Although we supervise our model with 2D SDF labels, our symmetric difference loss prevents the neural field from learning each 2D SDF beyond its interior/exterior classification. For this reason, the optimization could be driven using indicator function labels (\eg `1' for exterior and `-1' for interior). In \cref{fig:ablation_dog} we present the result of doing so on the Balloon Dog scene (denoted as ``No 2D SDF Labels"). We find that this leaves the predictable laddering artifacts along each contour. We attribute this to producing a discontinuous loss surface at the classification boundary. Conversely, our use of an L2 loss on 2D distance labels provides a smooth guide to this boundary.

\begin{table}[t]
    \centering
    \resizebox{0.65\columnwidth}{!}{
    \begin{tabular}{l|cc}
        \toprule
        & \multicolumn{2}{c}{\textbf{Elephant}} \\
        \textbf{Method} & \textbf{CD} ($\downarrow$) & \textbf{HD} ($\downarrow$) \\
        \midrule
         CrossSDF (full model) & \textbf{0.58} & \textbf{7.1} \\
          Standard Hash-grid & 1.51 & 15.0 \\
            \bottomrule
    \end{tabular}
    }
    \vspace{-6pt}
    \caption{Ablation of \method with all the  components removed on the Elephant scene using aligned planes.}
    \label{tab:elephant ablation}
    \vspace{-4pt}
\end{table}

\paragraph{Standard Hash-grid.} 

In addition to the baselines reported in the main paper, we conduct a further ablation by removing all of the key components of our \method approach; \ie the adaptive encoding, Fourier features, and symmetric difference loss. 
We denote this as the `Standard Hash-grid' baseline. 
The quantitative results for the Elephant scene are presented in \cref{tab:elephant ablation}. Removing these components leads to a degradation in performance, with the full model achieving the lowest CD and HD values.

\subsection{Supplementary Video} 
In the video, available on our website, we provide comparison of a reconstruction from our approach compared to the baselines.  
We also visualize the optimization process on a different scene.

\section{Implementation Details} 

\label{sec:implmentation_details} 
\begin{figure*}[htbp]  
\centering
\includegraphics[width=1\textwidth]{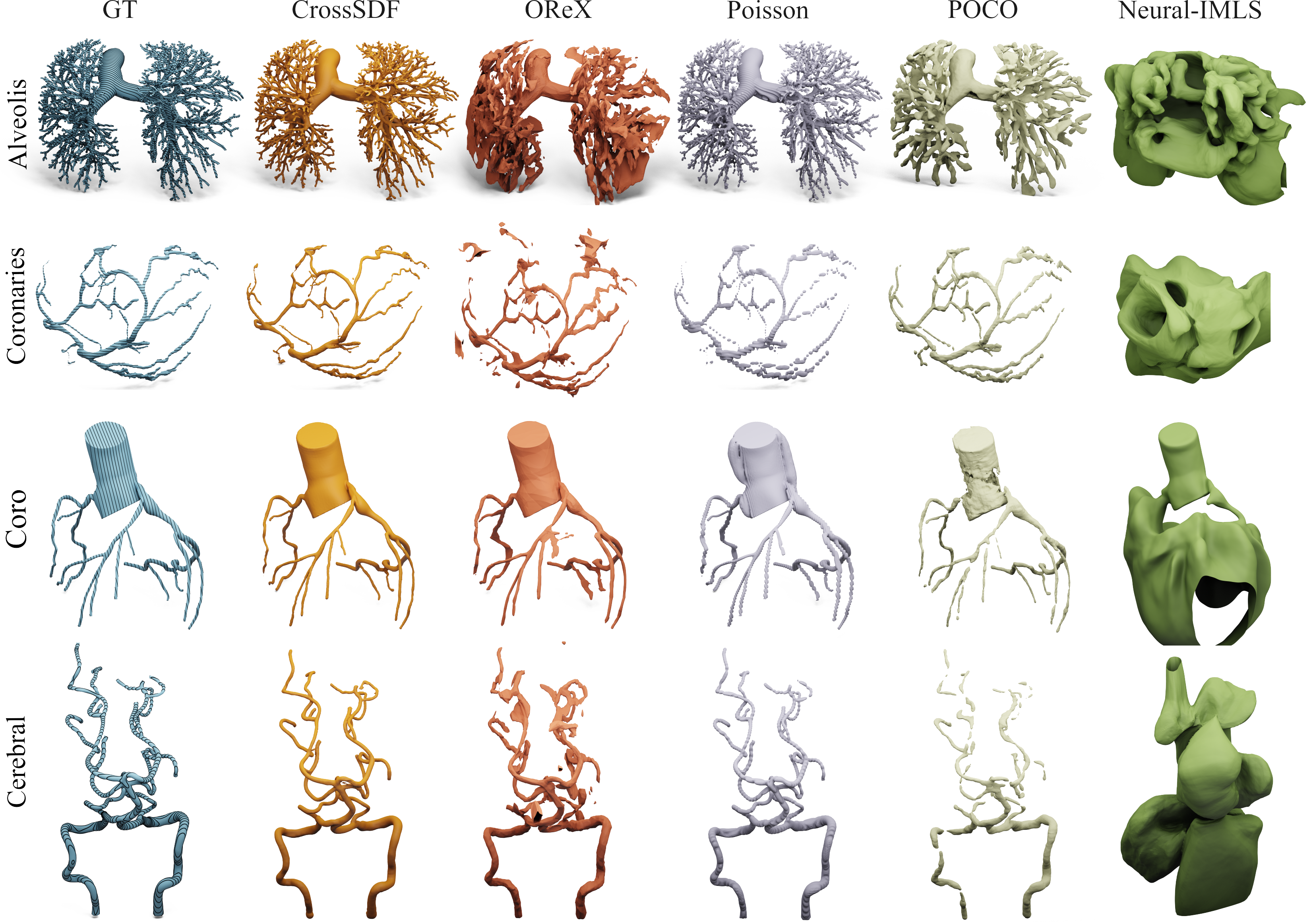}
\vspace{-10pt}  
\caption{Additional qualitative results of thin synthetic data in the aligned setting. We compare our model with existing methods. 
}
\label{fig:supp_thin}
\vspace{5pt}  
\end{figure*}

\begin{figure*}[htbp]  
\centering
\includegraphics[width=1\textwidth]{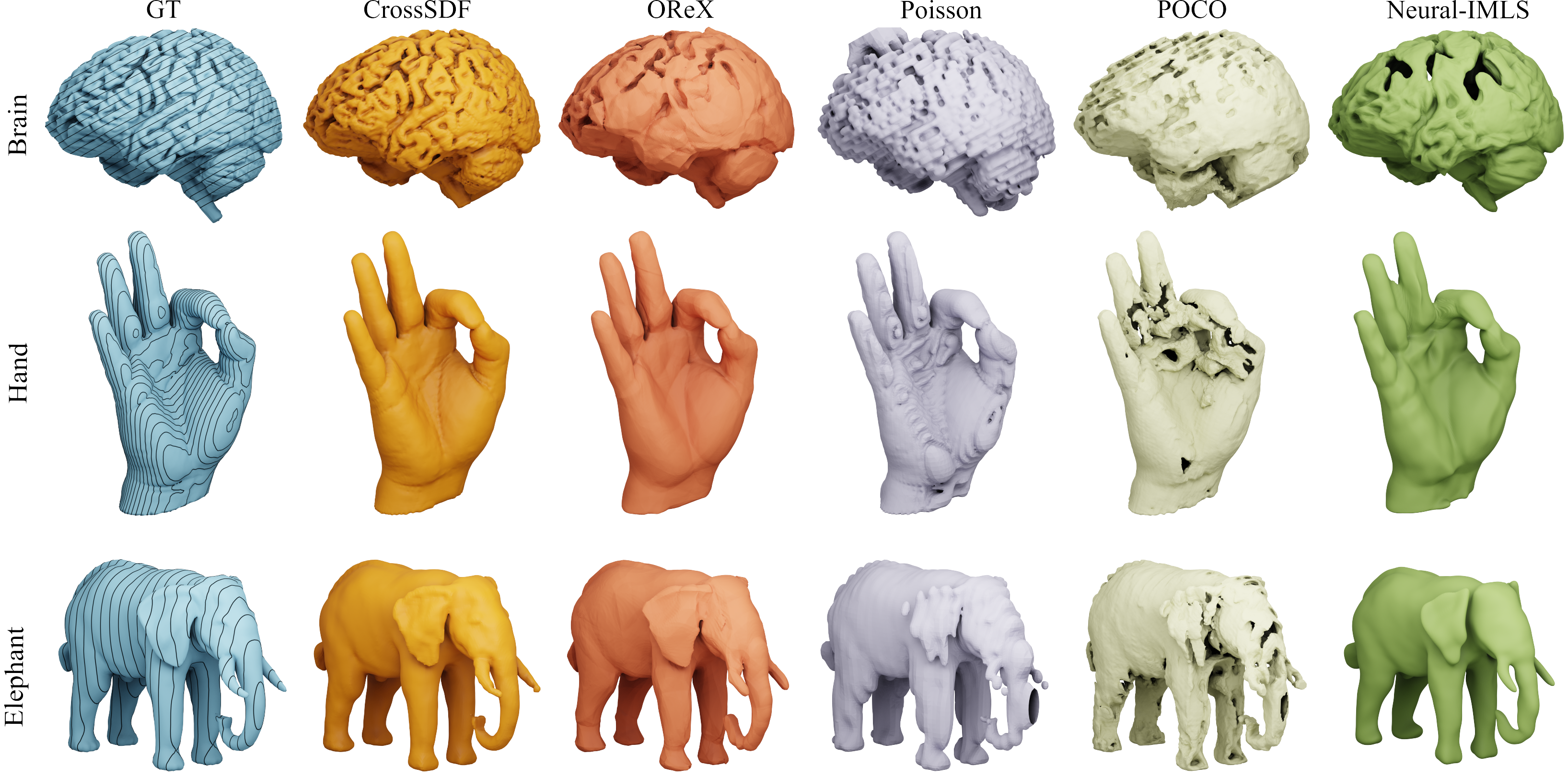}
\vspace{-14pt}  
\caption{Additional qualitative results of different methods on the thick synthetic dataset, for aligned slices. We compare our model with existing methods.} 
\label{fig:supp_thick}
\vspace{5pt}  
\end{figure*}

We implemented our solution in PyTorch and used the accelerated tiny-cuda-nn \cite{tiny-cuda-nn} implementation for our hash-encoding module. 
All scenes were trained using a single 24GB NVIDIA RTX 4090.
We train our network for a total of 500 epochs. The running time is commensurate with the complexity of the scene, taking as little as $~20$ minutes in the simplest cases (\eg Elephant in \cref{fig:supp_thick}) to $~5$ hours in the most complex scenes (\eg Alveolis in \cref{fig:supp_thin}). Our method does not have to create a (cubic-complexity) arrangement of cross-sectional planes, unlike \cite{bermano_arbitrary, TaoJu}, and thus we scale better with the input complexity, only depending on the number of samples. Our network hyper-parameters are not tuned for each dataset. 
We use the same weights for losses and hash encoding resolution for all the datasets displayed. The mesh is extracted using marching cubes at $512^3$ at the end of training (similarly for competing methods). 

\subsection{Model Architecture} 
In \cref{tab:hyper_params} we provide our hyper-parameters and network specifications. We use geometric initialization \cite{yariv2020multiview} to start the model as a sphere at the beginning of training. The neural networks are trained using the ADAM optimizer. The learning rate is initialized to $5\times10^{-4}$, and reduced by a factor of $0.9$ every $10$ epochs.

\begin{table}[h]
\centering
\resizebox{0.9\columnwidth}{!}{
\begin{tabular}{l|c}
\hline
\textbf{Hyper-parameter} & \textbf{Value} \\ \hline
\textbf{SDF Network Parameters} & \\ \hline
$M_{\text{SDF}}$ hidden layers  &  1 \\ \hline
$M_{\text{SDF}}$ hidden layer width &  256\\ \hline
$M_{\text{hash}}$ hidden layers  &  1 \\ \hline
$M_{\text{hash}}$ hidden layer width &  128\\ \hline
$M_{\text{RFF}}$ hidden layers  &  1 \\ \hline
$M_{\text{RFF}}$ hidden layer width &  128\\ \hline
Constant for RFF scaling $\alpha$ & 0.1\\ \hline
Softplus Activation $\beta$ & 100 \\ \hline

\textbf{Hash-Grid and Fourier Features} & \\ \hline
Hash-grid levels &  16 \\ \hline
Minimum hash-grid resolution &  $2^5$ \\ \hline
Maximum hash-grid resolution &  $2^{10}$\\ \hline
Hash-grid feature dimension & 4 \\ \hline
Hash dictionary size & $2^{22}$\\ \hline
Gaussian distribution variance for RFF &  1.0 \\ \hline 

\textbf{Sampling and Regularization}   & \\ \hline
Batch Size & $2^{17}$ \\ \hline
Minimum Surface $\beta$ & 100 \\ \hline
Threshold Samples Per Contour & 50 \\ \hline
$\lambda_{\text{eik}}$ Eikonal reg.~weight  &  $1\times10^{-3}$ \\ \hline
$\lambda_{\text{min}}$ minimum surface reg.~weight & $5\times10^{-2}$ \\ \hline
Weight decay weight & $2\times10^{-3}$ \\ \hline

\end{tabular}
}
\vspace{-5pt}
\caption{Hyperparameters and network specifications.
}
\label{tab:hyper_params}
\end{table}

\subsection{Data Pre-Processing}
\paragraph{Contouring.} For our synthetic dataset, we extract contours (as polylines) by taking the mesh and cutting plane parameters, and use this to compute a set of contours that results from cutting the mesh with the plane (\eg each vertex has two connecting edges). For medical  CT scans, human-annotated segmentations come in pixelized binary mask format, with slices along one axis. For each mask we generate a contour by running marching squares at 512 resolution.

\paragraph{Sampling.}
For the planar sampling $\Omega_{\text{pl}}$, we employ a combination of uniform sampling, on-contour sampling, and fixed-radius sampling, as described below:

\begin{itemize} 

\item \textbf{On-Contour Sampling:} ($\Omega_{\text{on}}$): We uniformly sample 25 points along each edge of the contour. 

\item \textbf{Fixed-Radius Sampling:} For each on-contour sample $\mathbf{x} \in \Omega_{\text{on}}$, we take two additional samples, each located a fixed distance $\epsilon$ from the edge in the perpendicular direction. One sample is positioned outward, perpendicular to the contour, while the other is positioned inward. 

\item \textbf{Uniform Sampling:} In each slicing plane, we perform uniform sampling, generating 10,000 samples per plane. 

\item \textbf{Adaptive Contour Sampling:} We sample a bounding box around each contour until at least 50 interior samples are gathered. This ensures thin structures with low cross-sectional area are captured.
\end{itemize}

Following a similar approach to OReX, we compute labels for the newly generated samples at predetermined intervals of 0, 50, 100, 200, and 300 epochs as part of a pre-processing step. During each re-sampling phase, the fixed radius $\epsilon$ is gradually reduced over time with values $\epsilon = 2^{-5}, 2^{-6}, 2^{-7}, 2^{-8}, 2^{-8}$. The label pre-processing step is efficient, taking less than three minutes for all tested scenes.

\subsection{Baselines}
For the point cloud reconstruction baseline methods, we use the on-contour samples $\Omega_{\text{on}}$ as a dense point cloud along each contour. For POCO \cite{boulch2022poco}, we use the model pre-trained on the ABC 10K dataset \cite{Koch_2019_CVPR}, which produced the best results relative to the alternative pre-trained models available.

\end{document}